\documentclass[journal]{IEEEtran}
\usepackage[switch]{lineno}
\usepackage{cite}
\usepackage{subfigure}
\usepackage{amsmath,amssymb,amsfonts}
\usepackage[noend]{algpseudocode}
\usepackage{algorithmicx,algorithm}
\usepackage{algpseudocode}
\usepackage{multirow}
\usepackage{amsmath}
\usepackage{array}
\usepackage{graphicx}
\usepackage{caption}
\usepackage{subcaption}
\usepackage{overpic}
\usepackage{textcomp}
\usepackage{xcolor}
\usepackage{stfloats}
\usepackage{textcomp}
\usepackage[colorlinks,
            linkcolor=blue,
            anchorcolor=black,
            citecolor=black]{hyperref}

\newcolumntype{L}[1]{>{\raggedright\let\newline\\\arraybackslash\hspace{0pt}}m{#1}}
\newcolumntype{C}[1]{>{\centering\let\newline\\\arraybackslash\hspace{0pt}}m{#1}}
\newcolumntype{R}[1]{>{\raggedleft\let\newline\\\arraybackslash\hspace{0pt}}m{#1}}

\begin{document}
% \pagewiselinenumbers 
% \switchlinenumbers 

\title{Heterogeneous Graph Reinforcement Learning for Dependency-aware Multi-task Allocation in Spatial Crowdsourcing}

\author{
        anonymous\IEEEmembership{}
}

\author{
        Yong~Zhao,\IEEEmembership{}
        Zhengqiu~Zhu,\IEEEmembership{}
        Chen~Gao,\IEEEmembership{}
        En~Wang,\IEEEmembership{~Member, IEEE}
        Jincai~Huang,\IEEEmembership{}
        and Fei-Yue~Wang,\IEEEmembership{~Fellow,~IEEE}

\thanks{This study is supported by Youth Independent Innovation Foundation of NUDT (ZK-2023-21) and the National Natural Science Foundation of China (62202477, 62173337, 21808181, 72071207). }
\thanks{Yong Zhao, Zhengqiu Zhu, and Jincai Huang are with the College of Systems Engineering, National University of Defense Technology, Changsha 410073, Hunan Province, China. 
(e-mail:~zhaoyong15@nudt.edu.cn;~zhuzhengqiu12@nudt.edu.cn;~huangjincai@nudt.edu.cn).}
\thanks{Chen Gao is with the BNRist, Tsinghua University, Beijing 100084, China. 
(e-mail:~ chgao96@gmail.com).}
\thanks{En Wang is with the College of Computer Science and Technology, Jilin University, Changchun 130012, China, and also with the Key Laboratory of Symbolic Computation and Knowledge Engineering of Ministry of Education, Jilin University, Changchun 130012, China. 
(e-mail:~ wangen@jlu.edu.cn).}
\thanks{Fei-Yue Wang is with the State Key Laboratory for Management and Control of Complex Systems, Institute of Automation, Chinese Academy of Sciences, Beijing 100190, China. 
(e-mail:~feiyue@ieee.org).}
}

% \markboth{IEEE TRANSACTIONS ON SYSTEMS, MAN, AND CYBERNETICS: SYSTEMS,~Vol.~X, No.~X, MAY~2024}%
\markboth{IEEE TRANSACTIONS ON X,~Vol.~X, No.~X, JUNE~2024}
{Shell \MakeLowercase{\textit{et al.}}: A Sample Article Using IEEEtran.cls for IEEE Journals}

\maketitle
% \linenumbers % 开启行号

\begin{abstract}

Spatial Crowdsourcing (SC) is gaining traction in both academia and industry, with tasks on SC platforms becoming increasingly complex and requiring collaboration among workers with diverse skills. Recent research works address complex tasks by dividing them into subtasks with dependencies and assigning them to suitable workers. However, the dependencies among subtasks and their heterogeneous skill requirements, as well as the need for efficient utilization of workers' limited work time in the multi-task allocation mode, pose challenges in achieving an optimal task allocation scheme. Therefore, this paper formally investigates the problem of Dependency-aware Multi-task Allocation (DMA) and presents a well-designed framework to solve it, known as Heterogeneous Graph Reinforcement Learning-based Task Allocation (HGRL-TA). To address the challenges associated with representing and embedding diverse problem instances to ensure robust generalization, we propose a multi-relation graph model and a Compound-path-based Heterogeneous Graph Attention Network (CHANet) for effectively representing and capturing intricate relations among tasks and workers, as well as providing embedding of problem state. The task allocation decision is determined sequentially by a policy network, which undergoes simultaneous training with CHANet using the proximal policy optimization algorithm. Extensive experiment results demonstrate the effectiveness and generality of the proposed HGRL-TA in solving the DMA problem, leading to average profits that is 21.78\% higher than those achieved using the metaheuristic methods.

\end{abstract}


\section{Introduction}

\IEEEPARstart{T}{he} advent and widespread adoption of smart devices and 5G technology have facilitated the extensive development of Spatial Crowdsourcing (SC) \cite{tong2020spatial,liu2020crowdos,wang2020compact,zhang2022online,xu2022incentive,wang2023ropriv}, attracting attention from both academic and industry. Different from traditional crowdsourcing \cite{estelles2012towards}, SC requires workers to arrive at the specific spatial and temporal location to participate and complete tasks. To facilitate the SC campaign, tasks are usually collected and allocated periodically by the platforms. These tasks can range from simple and straightforward activities like delivering food \cite{liu2018foodnet,li2022auction} or capturing images of landmarks \cite{wang2014smartphoto,wang2023leto}, to more intricate campaigns that require the collaborative efforts of workers with diverse skills, such as holding a wedding \cite{liu2022multi}, repairing the house\cite{yao2022online}, and refereeing a sports game \cite{ni2020task}. 

\begin{figure}
	\centering
	\includegraphics[width=\linewidth]{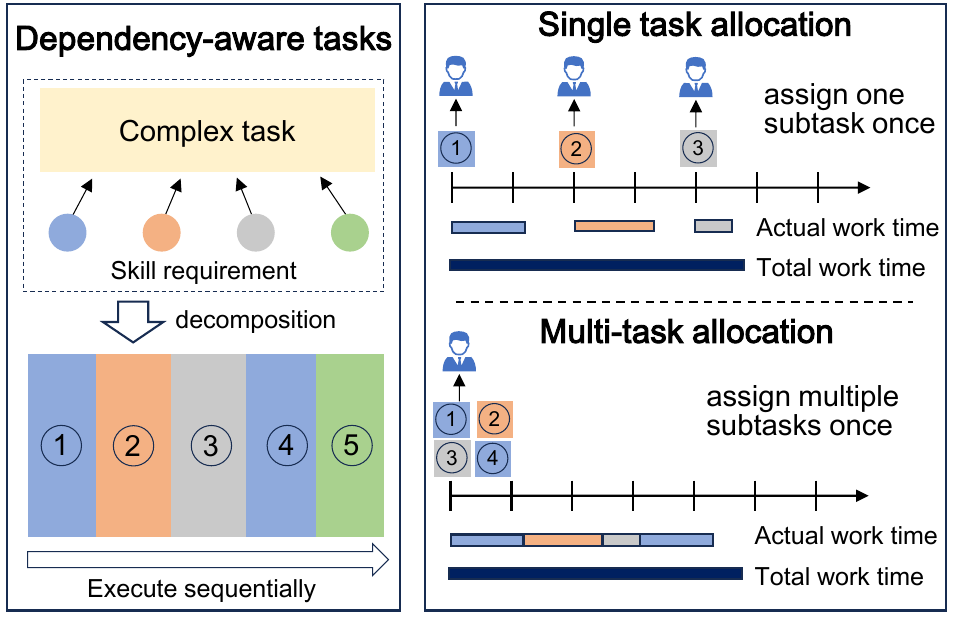}
	\caption{The diagram of related definitional items in this paper.}
	\label{fig:intro}
\end{figure}

To tackle the problem of complex spatial task allocation, several studies concentrate on matching a group of workers possessing the requisite skills essential for complex tasks \cite{cheng2016task,gao2016top}. Nevertheless, workers are often scarce, making it challenging to assemble a group of workers that fulfills all the skills needed for complex tasks. Therefore, the decomposition-based methods have been utilized in some research, wherein complex tasks are decomposed into multiple subtasks (or stages) with different skill requirements. These subtasks are then assigned independently to suitable workers\cite{ni2020task}, as shown in Fig. \ref{fig:intro}. For example, \textit{the complex task like house repairing can be divided into repairing the main body, installing electronic components and pipe systems, tiling the floor, and finally cleaning the rooms} \cite{yao2022online}.  It is noteworthy that in this case, the subtask of repairing the main body must be completed prior to any other subtasks, while the subtask of cleaning the rooms typically follows the completion of all other subtasks. This exemplifies a common reality where dependency constraints often arise among subtasks resulting from complex task decomposition. Therefore, the Dependency-aware Task Allocation (DTA) problem has gained increasing attention. In this problem, dependency relationships among subtasks are formulated as constraints and recent works have provided solutions for both the offline and online versions of this problem\cite{ni2020task,yao2022online,liu2022multi}. However, these works predominantly employ a single-task allocation mode, in which each worker is assigned only one task once. Under this mode, if a worker desires to undertake multiple tasks for increased profitability, they are required to wait and engage with the platform over several rounds of allocations.  Given that both tasks and workers are typically time-sensitive, the single-task allocation may result in suboptimal utilization of workers' limited work time, as shown in Fig. \ref{fig:intro}.

In contrast, multi-task allocation have been investigated by many studies as a more effective model, where multiple tasks are assigned to workers within each time slot, allowing for high utilization of workers' limited work time\cite{li2019multi}, as shown in Fig. \ref{fig:intro}. Nonetheless, successful multi-task allocation requires not only matching tasks with workers but also considering how to optimize worker efficiency when completing multiple assigned tasks \cite{liu2016taskme, lu2023incentivizing,shen2023heterogeneous}. Specifically,  it is necessary to determine a path for each worker, starting from their initial position and sequentially connecting the assigned tasks, while also establishing the start time for each task along the path.

Therefore, this paper aims to investigate the Dependency-aware Multi-task Allocation (DMA) problem. Specifically, we consider the dependencies among multiple subtasks within a complex task and perform the multi-task allocation to maximize overall benefits. The DMA problem is a typical NP-hard problem (as proven in this paper). Previous studies have proposed numerous heuristic \cite{han2021online,zhu2020cost,zhu2022crowd,zhao2023cost} and metaheuristic \cite{tao2020profit,li2019multi} approaches to obtain approximate solutions for NP-hard task allocation problems. However, these methods often lack stability and guaranteed performance across various structural instances or require significant time consumption. Additionally, the recent advancements in Heterogeneous Graph Reinforcement Learning (HGRL) method for solving combinatorial optimization problems\cite{peng2021graph} have motivated us to apply this method to tackle DMA problem. HGRL has demonstrated its ability to acquire knowledge from a set of problem instances (i.e. a training dataset) and subsequently apply that acquired knowledge to solve other similar instances \cite{song2022flexible,zhao2024application}. As a result, this approach showcases a notable level of generalization and typically offers satisfactory solving speeds. 

However, to achieve a robust generalization capability of the HGRL method on DMA problem across diverse instances, three challenges need to be addressed:

\begin{itemize}

\item The DMA problem involves not only matching workers to subtasks, but also sequencing their completion.  Additionally, the presence of dependency relations among subtasks emphasizes the importance of determining their completion order as it directly affects feasibility.

\item The structure and scale of DMA problem instances vary significantly, with different numbers of workers and subtasks. Besides, workers and subtasks have diverse characteristics and involve complex relations such as skill matching and dependencies among them. Representing this information concisely in a unified model poses a challenge.

\item  The problem state embedding presents a challenge in aggregating node and edge features based on the graph structure to obtain high-dimensional representations, as well as utilizing these representations for creating suitable graph embeddings for input into the policy network.

\end{itemize}

To address the DMA problem effectively in light of the aforementioned challenges, this paper proposes a well-designed framework called HGRL-TA. It tackles the DMA problem by treating it as a Markov Decision Process (MDP), where a subtask is sequentially assigned to a compatible worker until all feasible assignments are exhausted. Specifically, a multi-relation graph is devised to represent the problem state, capturing the diverse characteristics of workers and tasks, as well as the three types of relations among them. Besides, the Compound-path-based Heterogeneous Graph Attention Network (CHANet) is proposed to encode the graph, wherein node embedding is initially extracted and then utilized to derive embeddings for both states and actions through graph embedding.  Subsequently, these embeddings are fed into the policy network to make decisions while concurrently training both the policy network and embedding network using Proximal Policy Optimization (PPO) method.

To summarize, this paper presents the follow contributions:

\begin{itemize}

\item  The DMA problem is formally defined and formulated in this paper, which is proved as NP-hard. To address it, a graph-based solution is proposed, involving sequential subtask allocation that demonstrates robust generality across diverse problem instances.

\item The multi-relation graph model is elaborately designed to provide a concise and unified representation for diverse problem instances, with workers and subtasks depicted as nodes characterized by various static and dynamic features.  The edges connecting these nodes reflect distinct semantic relations including dependency, skill matching, and spatial relationships.

\item The CHANet is proposed as an embedding network that utilizes a compound-based approach to aggregate comprehensive information between two nodes. It differs from traditional meta-path-based Heterogeneous Graph Neural Networks (HGNNs) and demonstrates effective performance on the DMA problem, where decision-making is related to each node's embedding.

\item Extensive experiments are conducted to investigate three research questions. The results showcase the effectiveness and robust generality capacity of the proposed method, resulting in average profits that surpass those attained through metaheuristic approaches by 21.78\%.

\end{itemize}

The remainder of this paper is organized as follows: Section II provides a comprehensive review of the related work.  Section III presents the system model and problem formulation.  In Section IV, we introduce the proposed method.  The experimental setting and results are presented in Sections V.  Finally, in Section VI, we provide the conclusion of this paper.
\section{Related Works}

This section provides a comprehensive review of the relevant literature in this paper, primarily encompassing task allocation in spatial crowdsourcing and graph reinforcement learning for spatial crowdsourcing.

\subsection{Task Allocation in Spatial Crowdsourcing}

The task allocation problem, which involves recruiting a group of workers to complete a set of tasks, is one of the fundamental topic in SC. Previous studies have predominantly formulated this problem as an optimization problem, considering various optimization objectives and constraints \cite{guo2018task,wang2017multi,wang2022triple,bhatti2020approximation,zhang2019expertise}. The objectives encompass maximizing task completion, increasing profitability, and reducing costs, among others. Additionally, constraints are typically considered from perspectives such as time, cost, privacy protection, and the matching of worker skills with tasks. In recent years, some studies have begun to focus on task allocation problems with dependency constraints \cite{ni2020task,liu2022multi,yao2022online}. 
Ni et al. \cite{ni2020task} initially integrated multiple tasks with dependencies into an associative task set, which was subsequently treated as a complex task and assigned to a group of workers for completion. Besides, Liu et al. \cite{liu2022multi} investigated the Multi-Stage Complex Task Assignment (MSCTA) problem, which involves decomposing complex tasks into multiple dependent subtasks and allocating separate workers to each subtask. They devised a greedy algorithm and a game-theoretica algorithm for efficiently assigning the most profitable workers to these subtasks and achieved a provably approximate solution. Yao et al. \cite{yao2022online} studied an Online Dependent Task Allocation (ODTA) problem, taking into account spatial worker preferences. To maximize profits, they developed a threshold-based algorithm within the adversarial order model and achieved a near-optimal theoretical bound on the competitive ratio. However, these research has primarily focused on single-task allocation, which is suboptimal in scenarios where multiple tasks can be handled by a worker.

To fully utilize workers’ limited work time, the multi-task allocation problem has been investigated in recent years, leading researchers to propose diverse solutions from various perspectives. Zhang et al. \cite{zhang2021multi} proposed a improved evolutionary algorithm to solve the multi-task allocation problem to maximize the task completion, taking into account the daily routes of workers. Besides, Liu et al. \cite{liu2016taskme} proposed a new minimum cost maximum flow model to solve the multi-task allocation problem efficiently. The genetic algorithm is widely utilized for optimizing the allocation of multiple tasks to workers, aiming to maximize the platform's utility or enhance task quality of service \cite{li2019multi}. Moreover, an particle swarm optimization technique-based method was proposed by Estrada et al. \cite{estrada2017crowd}, which aims to maximize the ratio of aggregated quality of information to budget. Shen et al. \cite{shen2023heterogeneous} investigated a heterogeneous multi-project multi-task allocation problem based on the group collaboration mode, and proposed a multi-objective fireworks algorithm with dual-feedback ensemble learning framework to sovle the problem. However, most of these studies overlooked the dependencies among tasks.

In this paper, we take into account the dependencies among tasks (or subtasks) and multi-task allocation, while employing the HGRL-based approach to address the problem.

\subsection{Graph Reinforcement Learning for Spatial Crowdsourcing}

The effectiveness of Deep Reinforcement Learning (DRL) in solving sequential decision-making problems has been well-established, and in recent years it has been successfully integrated with Graph Neural Networks (GNNs) to address combinatorial optimization problems \cite{munikoti2023challenges,barrett2020exploratory,drori2020learning}. GNNs are deep learning models inherently designed to generalize over graphs of different sizes and structures, enabling the Graph Reinforcement Learning (GRL) approach to learn and generalize across diverse network topologies \cite{almasan2022deep}. For instance, Song et al.\cite{song2022flexible} demonstrated that the GRL method exhibits computational efficiency and outperforms traditional priority dispatching rules on the flexible job-shop scheduling problem, even when dealing with larger-scale instances and diverse properties not encountered during training. Moreover, Xu et al. \cite{xu2023intelligent} integrated a dedicatedly designed graph attention network into DRL to solve an multi-task allocation problem. Specifically, a homogeneous graph model is employed to represent the problem state, that is, worker and task nodes are represented by a same set of features, and edges between nodes are only characterized by the distance. However, the utilization of HGNNs becomes essential when the problem involves heterogeneous nodes or edges, as it enables a comprehensive representation of the problem state by incorporating rich semantics and structural information \cite{wang2022survey}.

The use of HGNNs has expanded to various tasks, including node classification, edge predictions, and analysis in domains like social networks, recommendation systems, and knowledge graph inference. \cite{yang2023simple,fang2023learning,huang2023egomuil,zhou2023predicting}. The meta-path-based methods are extensively employed in HGNNs to capture the structural information of the same semantic and subsequently integrate diverse semantic information. Initially, neighbor features are aggregated at the scope of each meta-path to generate semantic vectors, which are then fused to produce the final embedding vector \cite{wang2019heterogeneous}. The DMA problem addressed in this paper involves heterogeneous nodes and multiple relationships between these nodes, making it well-suited for resolution using HGNNs. However, to our knowledge, HGNN has not yet been applied to task allocation issues in spatial crowdsourcing. Therefore, this paper is the first to apply HGNN to the DMA problem and introduces a compound-path-based method, which demonstrates superior performance compared to the Meta-path-based method in addressing the DMA problem.

\section{System Model and Problem Formulation}
In this section, we present the main definitions and the formulation of the DMA problem. For clarity, the main notations are summarized in Table \ref{tab:I}.

\begin{table}
\centering
\caption{Main notations.}
\label{tab:I}
\renewcommand\arraystretch{1.2}
\resizebox{\linewidth}{!}{
\begin{tabular}{lp{5.5cm} l}
\hline
\hline
\textbf{Notations}                & \textbf{Explanations} \\
\hline
$u,p,v,U,P,V$       & the worker, task, subtask, and the sets for them          \\
$n,m,ml$              & the number of workers, tasks, and subtasks          \\
$l^u,\tau a,\tau w,\rho ,S{k^u}$       & the initial location, arrive time, work time, speed, and skill set of a worker          \\
$l^v,\tau s,\tau e,\tau p,b,S{k^v},D$       & the location, earliest start time, deadline, execution time, budget, required skill set, and dependency set of a subtask          \\
$v^u,V^u$       & the subtask assigned to the worker $u$, and the set for them          \\
\hline
\hline
\end{tabular}}
\end{table}

\subsection{System Model}
\textbf{Definition 1 (\textit{Heterogeneous worker}).} At a time slot, the spatial crowdsourcing platform collects a set of workers $U = \{ {u_1},{u_2}, \ldots{u_n}\} $. Each worker $u \in U$ can be characterized by a tuple of several attributes, i.e., $u = [ l^u,\tau a,\tau w,\rho ,S{k^u}] $. The ${l^u}$ denotes the initial location of the worker, while the $[\tau a,\tau a + \tau w]$ is the work time window. The $\rho$ represents the movement speed of the worker, while the $S{k^u}$ is a set of skill that the worker has mastered. 

\textbf{Definition 2 (\textit{Task and dependency-aware subtask}).} At a time slot, the spatial crowdsourcing platform collects a set of tasks $P = \{ {p_1},{p_2}, \ldots {p_m}\} $. Each task $p \in P$ consists of several subtasks (or stages), i.e., $p = \{ {v_1},{v_2}, \ldots {v_l}\} $ and each subtask $v \in p$ can be characterized as $v = [ l^v,\tau s,\tau e,\tau p,b,S{k^{v}},D] $. The ${l^{v}}$ is the location of the subtask, while the $[\tau s,\tau e]$ is the valid time window. To motivate workers to complete tasks, each subtask provides a budget $b$. In addition, the completion of each subtask requires a specific amount of time $\tau p$ and can only be achieved by a worker who possesses at least one of the required skills in $S{k^{v}}$. Besides, the subtasks contained in a task are dependent, implying that the execution of the subtask $v$ can only occur once all its dependent subtasks in the set $D$ have been completed. The dependency set of the subtask ${v_k}$ can be denoted as ${D(v_k)} = \{ {v_1},{v_2}, \ldots {v_{k - 1}}\} $. For clarity, the set of all subtasks is represented as $V = \{ {v_1},{v_2}, \ldots {v_{ml}}\} $, where $ml$ is the total number of subtasks.

\textbf{Definition 3 (\textit{Multi-task Allocation}).} The worker on the platform is willing to undertake multiple subtasks, provided that they possess the requisite skills and sufficient time. Therefore, for each worker $u \in U$, the platform will allocate several subtasks to he/she and the multi-task allocation is represented as $ <u,{V^u}> $, where ${V^u} = \{ v_1^u,v_2^u, \ldots v_o^u\} $ denotes the set of assigned subtasks and $o$ is the number of assigned subtasks. 
Upon receiving the multi-task allocations, workers will launch from their initial location and move to the assigned subtasks’ location to complete them sequentially. The start time of a subtask is denoted as $\tau b(v)$, which is not only determined by the time that the worker arrives the subtask, but also depends on the completion time of the subtasks in its dependency set. Specifically, the start time of the subtask $v_i^u \in {V^u}$ is calculated as:

\begin{equation}
\tau b(v_i^u) = \max \left( {{f_T}\left( {D(v_i^u)} \right),{f_U}\left( {u,v_i^u} \right)} \right)
\end{equation}
where the ${{f_T}\left( {D(v_i^u)} \right)}$ represents the latest completion time of the subtasks in the dependency set  $D(v_i^u)$, and the ${{f_U}\left( {u,v_i^u} \right)}$ denotes the arrive time of worker $u$ for the subtask $v_i^u$. They are calculated as follows:

\begin{equation}
{f_T}\left( {D(v_i^u)} \right) = \mathop {\max }\limits_{\forall v' \in D(v_i^u)} (\tau b(v') + \tau p(v'))
\end{equation}

\begin{equation}
{f_U}\left( {u,v_i^u} \right) = \left\{ \begin{matrix}
\tau a(u) + \rho (u) \cdot {f_D}\left( {l^u(u),l^v(v_i^u)} \right),i = 1 \hfill\\
\tau b(v_{i - 1}^u) + \tau p(v_{i - 1}^u) + \hfill\hfill\\
\quad\quad \rho (u) \cdot {f_D}\left( {l^v(v_{i - 1}^u),l^v(v_i^u)} \right),i > 1
\end{matrix}\right.
\label{eq3}
\end{equation}

In Eq. \ref{eq3}, the ${f_D}\left( \centerdot  \right)$ denotes the distance function between the two input locations, whereas in this paper, we employ the Euclidean distance in the function. When all assigned subtasks in ${V^u}$ are completed by the worker $u$, the platform will receive payment as a profit denoted by:

\begin{equation}
{f_P}\left( {{V^u}} \right) = \sum\limits_{{v^u} \in {V^u}} {b({v^u})}
\end{equation}

\subsection{Problem Formulation}

Based on the system model, the DMA problem is defined and formulated here.

\textbf{Definition 4 (\textit{DMA problem}).} Given a set of heterogeneous workers $U = \{ {u_1},{u_2}, \ldots {u_n}\} $ and a set of dependency-aware subtasks $V = \{ {v_1},{v_2}, \ldots {v_{ml}}\} $, the spatial crowdsourcing platform needs to implement the multi-task allocation for each worker as $ < u,{V^u} >$ , where ${V^u} \subseteq V$. The objective of the DMA problem is to maximize the overall profit of the platform, taking into account various constraints such as dependencies, skill matching, and time window limitation. Therefore, the DMA problem can be formulated as a combinatorial optimization problem:

\begin{equation}
\mathop {\max }\limits_{{V^u}:\forall u \in U} \sum\limits_{u \in U} {{f_P}({V^u})}
\label{egoal}
\end{equation}

\begin{equation}
\begin{aligned}
s.t.\quad \tau s({v^u}) \le &\tau b({v^u}) \le \tau e({v^u}) - \tau p({v^u}),\\
& \forall {v^u} \in {V^u},\forall u \in U
\end{aligned}
\label{c2}
\end{equation}

\begin{equation}
\begin{aligned}
\tau a(u) \le \tau b(v_1^u), &\tau b(v_o^u) + \tau p(v_o^u) \le \tau a(u) + \tau w(u), \\
& v_1^u,v_o^u \in {V^u},\forall u \in U
\end{aligned}
\label{c1}
\end{equation}

\begin{equation}
\begin{aligned}
\tau s(v') + \tau p(v') \le \tau s({v^u}),\\
\forall v' \in D({v^u}),\forall {v^u} \in {V^u},\forall u \in U
\end{aligned}
\label{c3}
\end{equation}

\begin{equation}
{S{k^u}(u) \cap S{k^{v}}({v^u}) \ne \emptyset },\quad \forall {v^u} \in {V^u},\forall u \in U
\label{c4}
\end{equation}

\begin{equation}
   {V^u} \cap {V^{u'}} = \emptyset, \quad  \forall u,u' \in U ,u \ne u'
\label{c5}
\end{equation}

Eq. \ref{egoal} is the goal of the problem. Eq. \ref{c1} and Eq. \ref{c2} demonstrate the time-window constraints of subtask and worker respectively. In Eq. \ref{c1}, the $v_1^u$ and $v_o^u$ denote the first and the last subtasks in the set ${V^u}$. Eq. \ref{c3} represents the dependency constraints among subtasks, while Eq. \ref{c4} ensures that the worker has the requisite skills to execute the assigned subtasks. Eq. \ref{c5} restricts that each subtask can be completed only once.
It is difficult to find the optimal solution of the DMA problem due to the large solution space. In fact, we can prove that the DMA problem is NP-hard.

\textbf{Theorem 1.} \textit{The DMA problem is NP-hard}.

\textbf{Proof.}We prove the theorem by reduction from an existing NP-hard problem, namely the Multiple Knapsack Problem (MKP), defined as follows: given a set $V = \{ {v_1},{v_2}, \ldots {v_n}\} $ of $n$ items and a set $U = \{ {u_1},{u_2}, \ldots {u_m}\} $ of $m$ knapsacks, where $m \le n$. Each item ${v_i} \in V$ has a weight ${w_i}$ and a utility ${p_i}$, while each knapsack ${u_j} \in U$ has a capacity ${c_j}$. The objective of the MKP is to optimize the allocation of items into knapsacks in order to maximize the overall utility. 

\begin{figure*}
	\centering
	\includegraphics[width=0.8\textwidth]{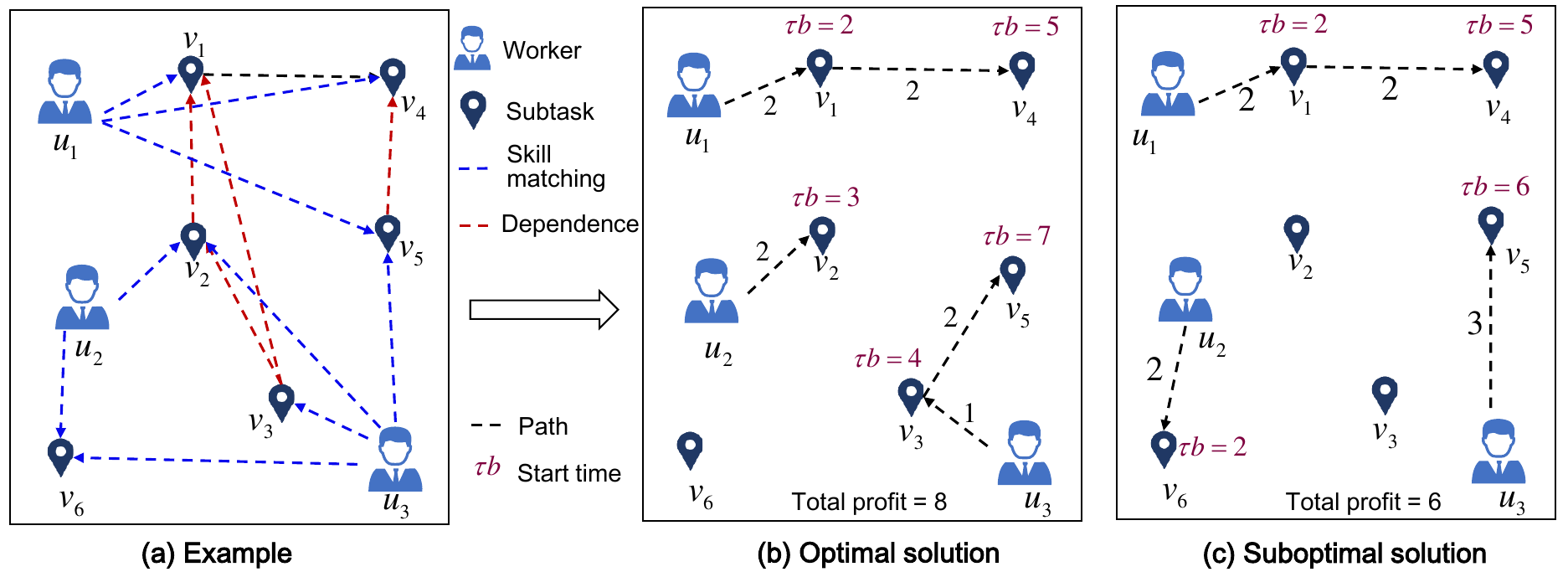}
	\caption{An illustrative instance of the DMA problem.}
	\label{fig:example}
\end{figure*}

Given an MKP, we can transform it to the simplified DMA problem within polynomial time. First, we assume that each task is composed of a single subtask and all workers possess the necessary skills to complete any task. Besides, the time required for a worker to transition between locations is considered as a constant value and the time-window constraint of subtasks are ignored. Then, we can associate the MKP and DMA by mapping the item to subtask and knapsack to worker. The budget of a subtask can be regarded as the utility of the item, while the travel time and execution time required for the subtask are associated with the weight of the item. Besides, the work time of the worker is associated to the capacity of the knapsack. For a simplified DMA problem, the goal is to allocation subtasks to worker such that the overall profit is maximized, which is same to the MKP. Thus, we can obtain that the MKP can be solved if and only if the simplified DMA problem can be solved. 

As demonstrated above, the MKP is as complex as the simplified DMA problem, which means the simplified MDA problem is also NP-hard. Therefore, the DMA problem can be proved as NP-hard.

\begin{table}
\centering
\caption{Details of tasks and workers.}
\label{tab:example}
\renewcommand\arraystretch{1.1}
\resizebox{\linewidth}{!}{
\begin{tabular}{cccccc}
\hline
\hline
Task                & Subtask & Skill & Dependency & Budget & Time Window \\ \hline
\multirow{3}{*}{$p_1$} & $v_1$  & $sk_1$ & $\emptyset $        & 1      & {[}0, 6{]}  \\
                    & $v_2$     & $sk_3,sk_4$  & $v_1$        & 1      & {[}0, 6{]}  \\
                    & $v_3$   & $sk_4$   & $v_1,v_2$        & 3      & {[}0, 6{]}  \\
\multirow{2}{*}{$p_2$} & $v_4$   & $sk_2$   & $\emptyset $    & 2      & {[}0, 8{]}  \\
                    & $v_5$   & $sk_1, sk_4$  & $v_4$       & 1      & {[}0, 8{]}  \\
$p_3$               & $v_6$   & $sk_3,sk_4$ & $\emptyset $  & 2      & {[}0, 3{]}  \\ \hline
\multicolumn{2}{c}{Worker}    & \multicolumn{3}{c}{Skill}   & Time Window \\ \hline
\multicolumn{2}{c}{$u_1$}        & \multicolumn{3}{c}{$sk_1, sk_2$}     & {[}0, 6{]}  \\
\multicolumn{2}{c}{$u_2$}        & \multicolumn{3}{c}{$sk_3$}     & {[}0, 4{]}  \\
\multicolumn{2}{c}{$u_3$}        & \multicolumn{3}{c}{$sk_2, sk_4$}     & {[}2, 7{]}  \\ \hline \hline
\end{tabular}}
\end{table}
\vspace{-5mm}

\subsection{Illustrative Instance}

We present an instance of the DMA problem that involves three complex tasks and three workers, with the three complex tasks decomposed into a total of six subtasks to be executed, as illustrated in Table \ref{tab:example}. The skill requirements for each subtask vary, and the workers possess diverse skills. Both workers and subtasks can involve multiple skills. The dependencies among subtasks can be determined based on the complex tasks they are associated with.

The data from the table is further transformed into a graph, as depicted in Fig. \ref{fig:example}.  In this graph, workers and subtasks are spatially distributed within a range, while the skill matching and dependency relationships are visually presented.  Additionally, Fig. \ref{fig:example} (b) and (c) provide two solutions for the instance. Each worker sequentially performs the assigned subtasks along the path indicated in the figure, with corresponding travel times displayed for each segment. The start time $\tau b$ for each subtask is also provided, and the execution time for each subtask is uniformly set to 1.

The optimal solution successfully completes five subtasks and attains a total profit of 8.  In contrast, the suboptimal solution only accomplishes four subtasks and yields a total profit of 6.  The key disparity between these two solutions lies in the timely completion of subtask $v_2$ by worker $u_2$ in (b), enabling worker $u_3$ to undertake the more lucrative subtask $v_3$. Additionally, it is noteworthy that the execution of subtasks is dependent on the completion of their preceding subtasks, regardless of whether a worker arrives at the subtask location early.  For example, in (b), worker $u_2$ reaches the location of subtask $v_2$ at timeslot 2, but $v_1$ has just started its execution. Consequently, $u_2$ must wait for the completion of $v_1$ before commencing $v_2$, resulting in an actual start time for $v_2$ as 3.

\section{Methodology}

\subsection{Heterogeneous Graph Reinforcement Learning-based Task Allocation}

\begin{figure*}[!htb]
	\centering
	\includegraphics[width=0.8\textwidth]{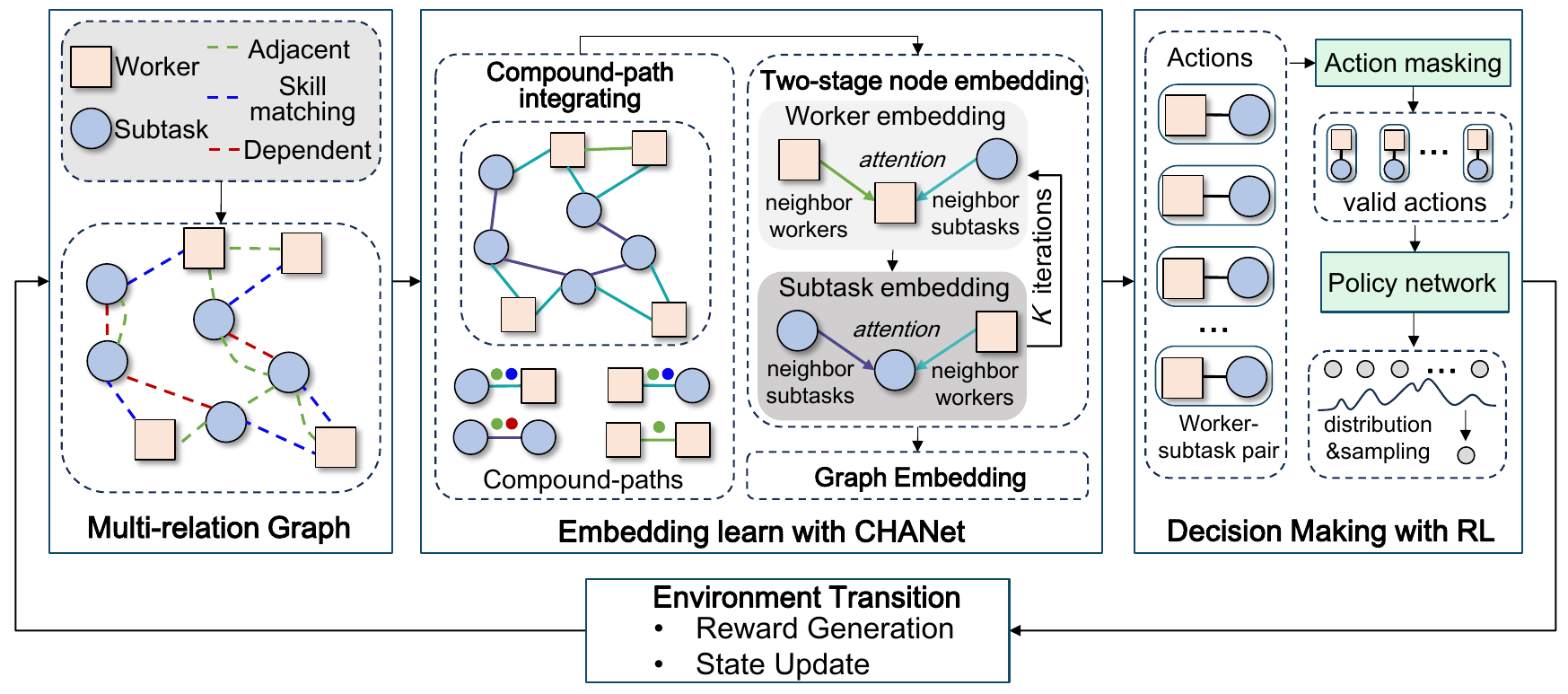}
	\caption{Overview of the proposed HGRL-TA.}
	\label{fig:framwork}
\end{figure*}

The overview of the proposed HGRL-TA framework is depicted in Fig. \ref{fig:framwork}. HGRL-TA employs three specific components to determine the optimal decision at each iteration. First, the problem state is formulated based on the multi-relation graph. Subsequently, the proposed CHANet is employed to encode the graph. After that, the embedding is subsequently fed into the policy network to obtain decisions, while both the policy network and embedding network are concurrently trained using the PPO method. The details are presented in this section.

The MDP mainly consists of five elements $(S,A,T,R,\pi )$. The state space and the action space of the MDP are denoted by $S$ and $A$, respectively. The transition function $T$ is utilized to facilitate the environment's transition to new states based on actions, while the reward function $R$ provides the reward from state transition. The policy $\pi$ is employed to optimize the selection of actions from the action space in order to maximize long-term rewards.

The state in each step consists of the conditions of all the subtasks and workers, as well as the multiple relations among them. The detailed representation of the state and the corresponding transition function will be described based on the multi-relation graph model in the next section. In each step, an action $a = (v,u) \in A$ is taken, which is defined as a pair of a feasible worker and an incomplete subtask. The action means the subtask is assigned to the worker. Given the constraints stated in Eq. \ref{c1}-\ref{c5}, not all subtask-worker pairs are valid at every step. Therefore, we employ an action masking process to filter out valid actions at each step, which also reduces the decision complexity. When the action $a = (v,u)$ is selected, the worker takes a certain amount of time to travel and execute the subtask, while the platform receives the budget $b(v)$ after the subtask is completed. Thus, we define the reward for taking the action as follows:

\begin{equation}
r = b(v) - \alpha \left( {\rho (u)\cdot{f_D}(lo{c^u},l^v(v)) + \tau p(v)} \right)
\label{eq:reward}
\end{equation}
where the $\rho (u)\cdot{f_D}(lo{c^u},l^v(v))$ represents the travel time of the worker to reach the subtask from his/her location $lo{c^u}$. The $lo{c^u}$ can refer to either the initial location of the worker or the location of the previous subtask that was executed by the worker. The $\alpha $ represents a constant that is utilized to account for the impact of the time consumed by worker on the reward. In addition, the policy network is utilized  to select an action from a probability distribution over the set of valid actions at each step.

\subsection{Multi-relation Graph}

The utilization of heterogeneous graph models is prevalent in representing combinatorial optimization problems, owing to their efficient representation of multiple node types and relations within the problem. Additionally, these models offer a unified representation framework that accommodates instances with varying numbers of nodes. For DMA problem, worker and subtask can be processed as heterogeneous nodes since they has different features, while the skill matching and dependency constraints can be modeled as the edge with according semantic relation. Besides, the spatial relation among workers and subtasks is also taken into consideration, as it directly impacts the worker's ability to access subtasks and the time required for movement. Consequently, a multi-relation graph model is established to represent the DMA problem, considering that there can be various types of relations between two nodes.

Theoretically, given an instance of the DMA problem that contains a set of workers $U = \{ {u_1},{u_2}, \ldots {u_n}\} $ and a set of subtasks $V = \{ {v_1},{v_2}, \ldots {v_{ml}}\} $, its state can be represented by a multi-relation graph $\cal G = \{ \cal U,\cal V,\cal E,\cal Z\}$. The ${\cal U}$ and ${\cal V} $ are the set of worker nodes and subtask nodes, respectively. The ${\cal E}$ is the set of edge with a mapping function $\psi :{\cal E} \to {\cal Z}$, that is, each edge $e \in {\cal E}$ is attached with a relation $z = \psi (e) \in {\cal Z}$. In this paper, three types of relations are considered, i.e., ${z_{\rm{sm}}}$: skill matching, ${z_{\rm{\rm{dp}}}}$: dependent, and ${z_{\rm{ad}}}$: adjacent. Therefore, there can be multiple edges between two nodes and the edges with different semantic relations exist independently.  

\subsubsection{Edge representation}

For the worker node $u$ and subtask node $v$, there is an edge with relation ${z_{\rm{sm}}}$ between them when the worker has the skill required by the subtask, i.e., $S{k^u}({u}) \cap S{k^{v}}({v}) \ne \emptyset $. As for the  edge with relation ${z_{\rm{dp}}}$, it exists between every subtask node that belongs to the same task. On the one hand, the execution of a subtask depends on the completion of its predecessor subtask. On the other hand, the successor subtasks of a subtask are related to the long-term reward of the execution of the subtask. Besides, the edge with relation ${z_{\rm{ad}}}$ can occur between any two nodes, as long as the distance between them is less than a certain threshold. 

\subsubsection{Node representation}

In the multi-relation graph model, each node can be represented by a set of raw features. Two types of heterogeneous nodes are represented using different features that encompass both static and dynamic characteristics. Specifically, we represent raw features of each worker node as ${x^u} = [x_1^u,x_2^u, \ldots x_8^u]$, where $x_1^u - x_6^u$ denotes the dynamic features of the worker, which are continuously updated throughout the sequential decision-making process. The $x_1^u$ denotes the available time for the worker, which is the moment when the worker joins in the platform or completes the predecessor subtask. The $x_2^u$ and  $x_3^u$ are the two-dimensional location. The $x_4^u$ represents the number of accessible subtasks that the worker can complete under various constraints, while $x_5^u$ denotes the total budget of these accessible subtasks. The $x_6^u$ is the profit that has obtained by the worker. The $x_7^u$ and $x_8^u$ are static features that represent the worker's speed and expire time, respectively.

Moreover, raw features of each subtask node can be represented as  ${x^v} = [x_1^{v},x_2^{v}, \ldots x_9^u]$, where $x_1^{v} - x_5^{v}$ denotes dynamic features. The $x_1^{v}$ is a Boolean variable used to represent the status of the subtask, where $x_1^{v}=1$ indicates that the subtask has been assigned, and $x_1^{v}=0$ indicates that the subtask has not been assigned. The $x_2^{v}$ represents the start time of the subtask, and in cases where the subtask has not been assigned, we assign it a value of a significantly large constant. The $x_3^{v}$ represents the number of workers capable of completing the subtasks and the $x_4^{v}$ indicates the number of incomplete subtasks within the dependency set of the subtask. The $x_5^{v}$ signifies the total budget for all incomplete subtasks that belong to the same task. This reflects the future profits associated with the subtask. The $x_6^{v} - x_9^{v}$ represents static features, specifically the two-dimensional location, budget, and deadline of the subtask.

\subsubsection{Graph Update}
The status and start time of subtask $v$ are updated when it is assigned to worker $u$. Subsequently, the available time of the worker is updated to the completion time of the subtask, and the location of the worker is set to match that of the subtask. The worker's ability to complete other subtasks is then evaluated based on their new location and available time, and the remaining dynamic features for both the worker and the subtask nodes are updated accordingly. Additionally, it is necessary to update the edges with relation ${z_{\rm{ad}}}$ among nodes due to the change in location.

\subsection{Compound-based Heterogeneous Graph Attention Network}

The CHANet customized for the DMA problem is proposed in this paper, and the embedding process based on CHANet is presented in Algorithm \ref{a1}. First, various meta-paths with distinct semantics are integrated into compound-paths during the pre-processing stage (Line 1-3). Subsequently, a two-stage node embedding process is employed to acquire the node embedding (Line 4-8). Finally, the state and action embeddings are derived from transforming the node embeddings through graph embedding (Line 9-10).

\begin{algorithm}[tp]
	\caption{State and Action Embedding based on CHANet.}
	\hspace*{0.02in} {\bf Input:} 
       \parbox[t]{\dimexpr\linewidth-\algorithmicindent}{Heterogeneous graph ${\cal G} = \{ {\cal U},{\cal V},{\cal E},{\cal Z}\} $, \\ embedding round $K$ } \\
	\hspace*{0.02in} {\bf Output:} %算法的结果输出
	Embedding vectors of state ${h^s}$ and action ${h^a}$
	\begin{algorithmic}[1]
         \State Obtain raw features of compound-paths 
         \State Obtain compound-path-based neighborhoods of each node
         \State Project raw features of nodes and compound-paths to obtain their initial vector
		\For {$k = 1:K$}
            \For {$n^u = \cal U$}
                \State Update $h_k^u$  according to Eq. \ref{eq.15} 
		      \EndFor
            \For {$n^v = \cal V$}
                \State Update $h_k^v$  according to Eq. \ref{eq.16} 
		      \EndFor
		\EndFor
        \State Construct the state embedding $h^s$ according to Eq. \ref{eq.17}
        \State Construct the action embedding $h^a$ according to Eq. \ref{eq.18}
        \State \textbf{Return}
	\end{algorithmic}
    \label{a1}
\end{algorithm}

\subsubsection{Compound-path integrating}

\begin{figure*}[!htb]
	\centering
	\includegraphics[width=0.8\textwidth]{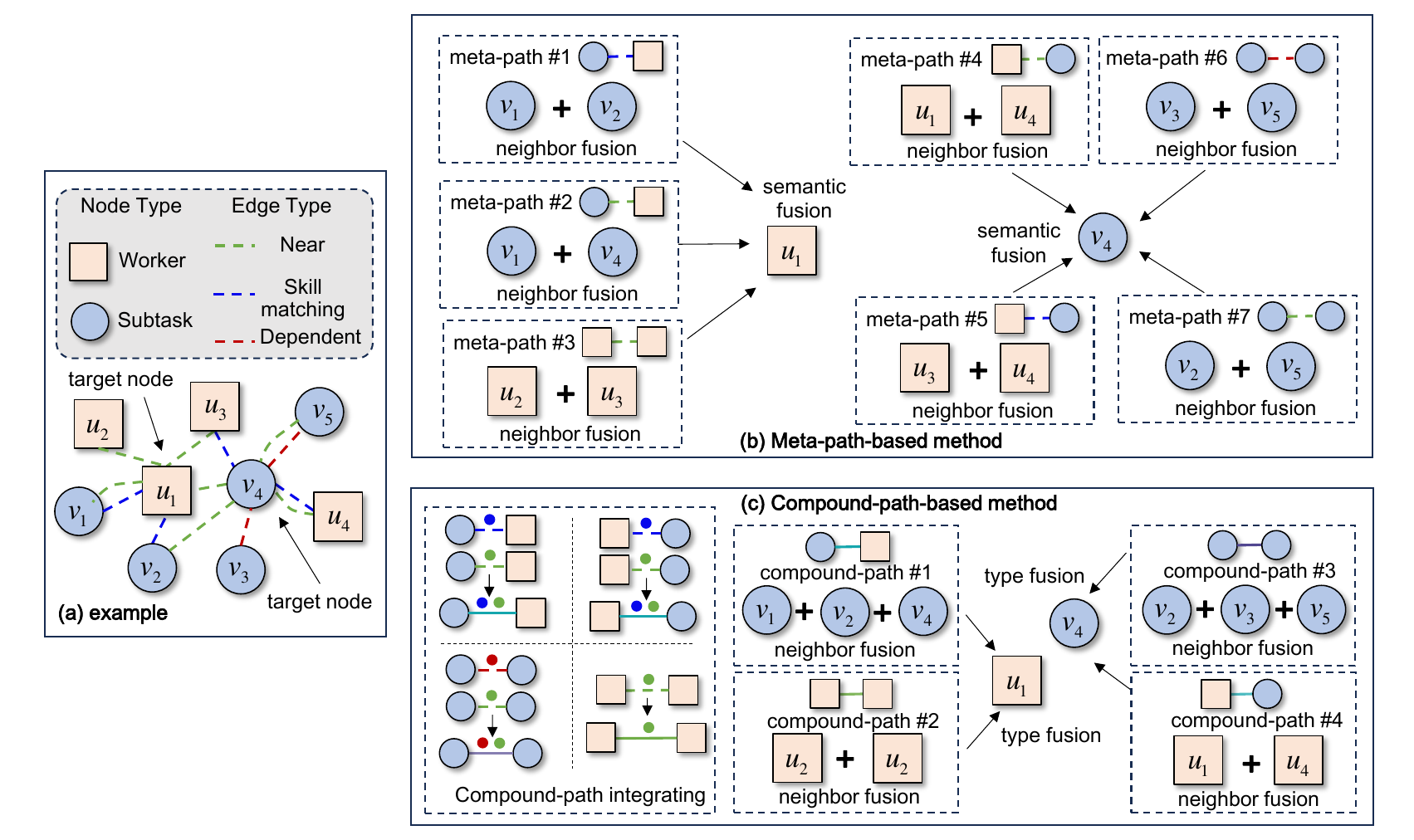}
	\caption{The illustration of the meta-path-based method and proposed compound-path-based method.}
	\label{fig:cp}
\end{figure*}

The meta-path-based method is a prominent category within the field of HGNNs, which finds extensive application in tasks such as node classification and link prediction. The meta-path-based approach utilizes meta-paths to establish high-level semantic connections between two nodes, subsequently performing the neighbor fusion to aggregate the neighbor information of the same semantic within each meta-path and then integrating diverse semantic information, as illustrated in Fig. \ref{fig:cp}(b). Theoretically, a meta-path defines a composite relation of several relations and nodes in the forms of ${{\cal P}} \buildrel \Delta \over = {n_t}\mathop  \leftarrow \limits^{{z_1}} {n_2}\mathop  \leftarrow \limits^{{z_2}}  \ldots \mathop  \leftarrow \limits^{{z_q}} {n_s}$ , which denotes a directed  $q$-hop relation from the source node $n_s$ to the target node $n_t$. Furthermore, the meta-path-based neighborhood is represented as  $N^{\cal P}$, which contains all nodes connected with the target node via the meta-path ${\cal P}$.

In this paper, the one-hop meta-paths encompass multiple relations, allowing for different one-hop meta-paths to connect two given nodes. For instance, a worker node $u$ may be connected with other nodes via three types of meta-path that are denoted as ${{\cal P}}_{\rm{ad}}^{uu} = {u}\mathop  \leftarrow \limits^{{z_{\rm{ad}}}} {u'}$, ${{\cal P}}_{\rm{ad}}^{uv} = {u}\mathop  \leftarrow \limits^{{z_{\rm{ad}}}} {v}$, and ${{\cal P}}_{\rm{sm}}^{uv} = {u}\mathop  \leftarrow \limits^{{z_{\rm{sm}}}} {v}$, where $u' \in \cal U$ and $v \in \cal V$. Similarly, a subtask node $v$ may be connected with other nodes via four types of meta-path that are denoted as ${{\cal P}}_{\rm{ad}}^{vu} = {v}\mathop  \leftarrow \limits^{{z_{\rm{ad}}}} {u}$, ${{\cal P}}_{\rm{ad}}^{vv} = {v}\mathop  \leftarrow \limits^{{z_{\rm{ad}}}} {v'}$, ${{\cal P}}_{\rm{sm}}^{vu} = {v}\mathop  \leftarrow \limits^{{z^{\rm{sm}}}} {u}$, and ${{\cal P}}_{\rm{dp}}^{vv} = {v}\mathop  \leftarrow \limits^{{z_{\rm{dp}}}} {v'}$, where $u \in \cal U$ and $v' \in \cal V$. Therefore, there are total seven types of meta-path considered in this paper.

The performance of downstream tasks, such as node classification and link prediction, typically relies solely on the final representation of the target node or edge that requires classification or prediction. Hence, in scenarios where multiple meta-paths exist between two nodes, employing a meta-path-based approach can facilitate more nuanced semantic information propagation, thereby leading to enhanced performance in these tasks. However, for task allocation problems such as the DMA, the downstream task relies on the representation of each node (i.e., selecting from all worker-subtask pairs). Therefore, instead of considering meta-paths as the fundamental unit of feature aggregation, it is more appropriate to regard nodes as the fundamental unit for aggregating features based on their overall relations. Consequently, diverse semantic information between two nodes represented by multiple meta-paths can be integrated into the compound-path, thereby facilitating the transfer of information based on the comprehensive node relation.

Therefore, this paper proposes a novel compound-path-based method, which facilitates the aggregation of structural and semantic information according to the node types. Specifically, the compound-path based method first integrates the diverse semantic relations between nodes to obtains several compound-paths. Subsequently, it performs neighbor fusion process that captures the compound information of nodes with the same node type within each compound-path-based neighborhood, then the type fusion is performed to consolidates information from different node types, as shown in Fig. \ref{fig:cp}(c).

Based on the types of source node and target node of the meta-paths, the seven meta paths can be integrated into four compound paths, denoted as ${{\cal C}}{{{\cal P}}^{uu}} = {u}\mathop  \leftarrow \limits^{c{z_{uu}}} {u'}$,  ${{\cal C}}{{{\cal P}}^{uv}} = {u}\mathop  \leftarrow \limits^{c{z_{uv}}} {v}$, ${{\cal C}}{{{\cal P}}^{vu}} = {v}\mathop  \leftarrow \limits^{c{z_{vu}}} {u}$, and ${{\cal C}}{{{\cal P}}^{vv}} = {v}\mathop  \leftarrow \limits^{c{z_{vv}}} {v'}$, where the $c{z_{uu}}$, $c{z_{uv}}$, $c{z_{vu}}$, and $c{z_{vv}}$ are the compound relations and their feature can be combined by the features on each meta-path. The feature of the relation ${z_{\rm{sm}}}$ and ${z_{\rm{dp}}}$ can be represented by a Boolean variable ${x^{\rm{sm}}}$ and ${x^{\rm{dp}}}$ respectively, where the feature equals 1 when the relation exists, otherwise, the feature equals 0. The feature of ${z_{\rm{ad}}}$ can be represented by the distance ${x^{\rm{ad}}}$ between two nodes, which provides more detailed spatial information. Therefore, the compound relations can be featured as ${x^{uu}} = [{x^{\rm{ad}}}]$, ${x^{uv}} = [{x^{\rm{ad}}},{x^{sk}}]$, ${x^{vu}} = [{x^{\rm{ad}}},{x^{sk}}]$, and ${x^{vv}} = [{x^{\rm{ad}}},{x^{sk}}]$ respectively. It is noteworthy that the compound-path employed in this paper is undirected, thus we have ${x^{uv}} = {x^{vu}}$

The compound-path-based neighborhoods can be obtained by merging the meta-path-based neighborhoods. For instance, the neighborhood based on compound-path ${{\cal C}}{{{\cal P}}^{uv}}$ is denoted as ${N_{uv}} = {N^{{{\cal P}}_{\rm{ad}}^{uv}}} \cap {N^{{{\cal P}}_{\rm{sm}}^{uv}}}$. The other three types of compound-path-based neighborhoods can be obtained as ${N_{uu}} = {N^{{{\cal P}}_{\rm{ad}}^{uu}}}$, ${N_{vu}} = {N^{{{\cal P}}_{\rm{ad}}^{vu}}} \cap {N^{{{\cal P}}_{\rm{sm}}^{vu}}}$, and  ${N_{vv}} = {N^{{{\cal P}}_{\rm{ad}}^{vv}}} \cap {N^{{{\cal P}}_{\rm{dp}}^{vv}}}$.

\subsubsection{Two-stage node embedding}
\begin{figure*}[!htb]
	\centering
	\includegraphics[width=0.7\textwidth]{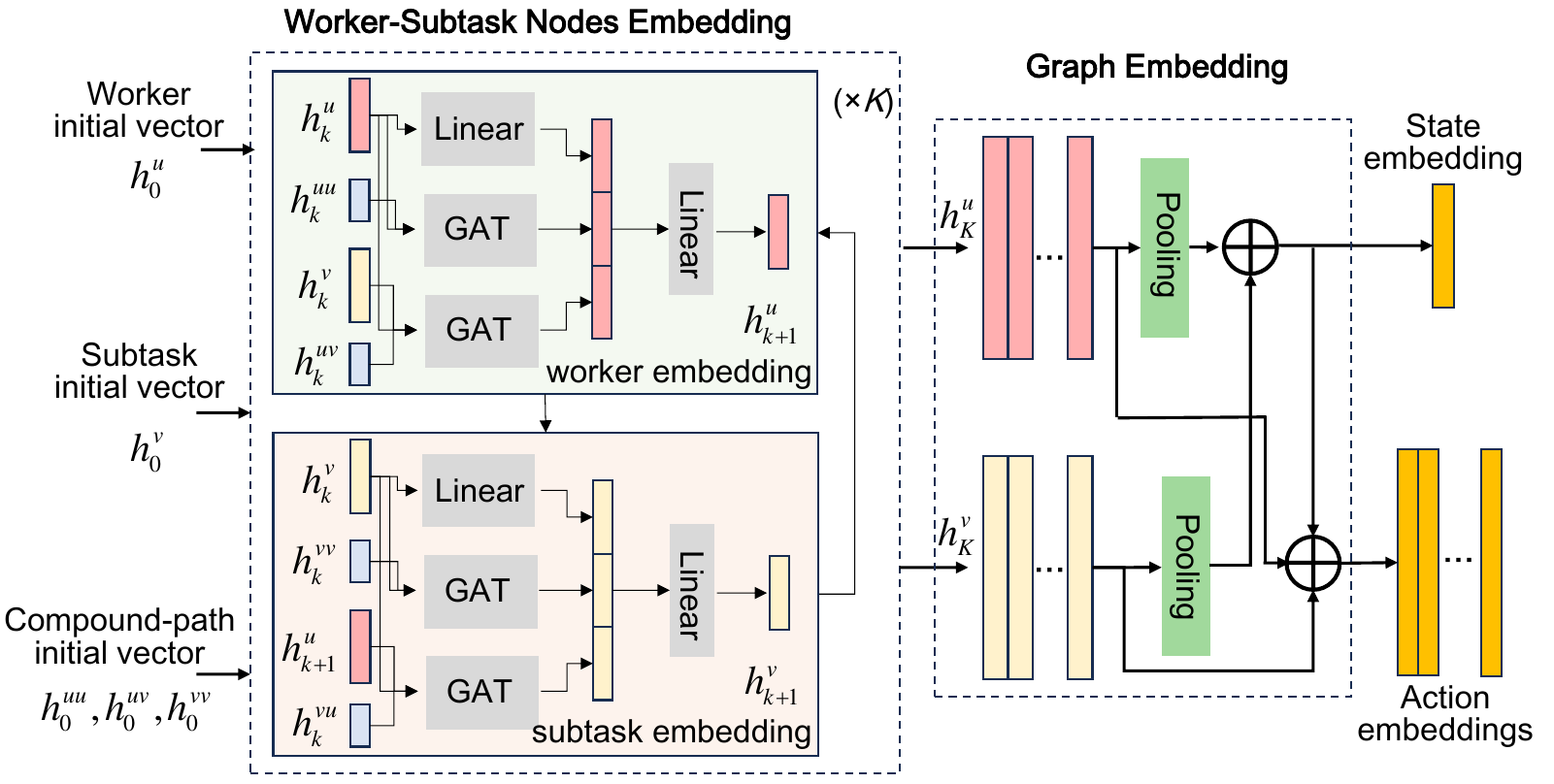}
	\caption{The architecture of CHANet.}
	\label{fig:chanet}
\end{figure*}

This paper adopts a two-stage node embedding method and utilizes the graph attention network to perform feature aggregation of neighboring nodes \cite{song2022flexible}. The architecture of CHANet is illustrated in Fig. \ref{fig:chanet}. The embedding process of worker nodes and subtask nodes is performed iteratively for $K$ rounds, similar to previous studies employing the two-stage embedding approach, in order to obtain the final vector representation $h_K$ for each node. The raw features of both nodes and edges are projected to the same dimension $\lambda $ through linear transformations to form the initial vector before embedding: $h_0^u = w_0^u{x^u}$, $h_0^v = w_0^v{x^v}$, ${h^{uu}} = {w^{uu}}{x^{uu}}$, ${h^{uv}} = {w^{uv}}{x^{uv}}$, ${h^{vv}} = {w^{vv}}{x^{vv}}$, where the $w_0^u \in {\mathbb {R}^{\lambda  \times 8}}$, $w_0^v \in {\mathbb {R}^{\lambda  \times 9}}$, $w_0^{uu} \in {\mathbb {R}^{\lambda}}$, and ${w^{uv}},{w^{vv}} \in {\mathbb {R}^{\lambda  \times 2}}$ are trainable weights of linear transformations.

For the worker node $u$ , it has two compound-path-based neighborhoods ${N_{uu}}$ and ${N_{uv}}$, comprising numerous nodes that may have different importance to $u$. Therefore, the Graph Attention neTwork (GAT) is employed to aggregate the neighbor features of the same node type within each compound-path, leveraging the attention mechanism to automatically learn their respective importance. Specifically, given the vector $h^t$ of target node, the vector $h^s$ of source node in the neighborhood $N_{ts}$, and the vector $h^{ts}$ of their compound relation, the attention coefficient between the two nodes can be calculated as:

\begin{equation}
{e_{ts}} = {\mathop{\rm elu}\nolimits} \left( {{{a}^{\mathop{\rm T}\nolimits} }\cdot{w_e}\cdot{\mathop{\rm cat}\nolimits} ({h^t},{h^s},{h^{ts}})} \right)
\end{equation}
where ${a} \in {\mathbb {R}^\lambda}$ and ${w_e} \in {\mathbb {R}^{3\lambda \times \lambda}}$ are the trainable weight of linear transformation. The ${\mathop{\rm elu}\nolimits} (\centerdot)$ is the activation function and ${\mathop{\rm cat}\nolimits} (\centerdot)$ is the short-hand of the concatenation function. Then the coefficients are normalized across the neighborhood using softmax function:

\begin{equation}
{\alpha _{ts}} = {{\exp ({e_{ts}})} \over {\sum\limits_{{n_{s'}} \in {N_{ts}}} {\exp ({e_{ts'}})} }}, \forall {n_s} \in {N_{ts}}
\end{equation}

Then, the neighbor feature from the neighborhood $N^{ts}$ can be aggregated based on the normalized attention coefficients and the vectors as:

\begin{equation}
{h^{{N_{ts}}}} = {\mathop{\rm elu}\nolimits}\left( {\sum\limits_{{n_s} \in {N_{ts}}} {{\alpha _{ts}}\cdot{w_{ts}}\cdot{\mathop{\rm cat}\nolimits} ({h^s},{h^{ts}})} } \right)
\end{equation}
where the ${w_{ts}} \in {\mathbb {R}^{2\lambda \times \lambda}}$ is the trainable weight of linear transformation.

Based on the attention mechanism, the vector of the worker node $u$ and subtask node $v$ at $k+1$ round can be obtained as:

\begin{equation}
\label{eq.15}
h_{k + 1}^u = {\mathop{\rm elu}\nolimits} ({w_u}\cdot{\mathop{\rm cat}\nolimits} (h_k^u,h_k^{{N_{uu}}},h_k^{{N_{uv}}}))
\end{equation}

\begin{equation}
\label{eq.16}
h_{k + 1}^v = {\mathop{\rm elu}\nolimits} ({w_v}\cdot{\mathop{\rm cat}\nolimits} (h_k^v,h_k^{{N_{vv}}},h_k^{{N_{vu}}}))
\end{equation}
where ${w_u},{w_v} \in {\mathbb {R}^{3\lambda  \times \lambda }}$ are the trainable weight of linear transformation. The $h_k^{{N_{uu}}}$, $h_k^{{N_{uv}}}$, $h_k^{{N_{vv}}}$, and $h_k^{{N_{vu}}}$ are the vectors of neighbor information gathered within the compound-path-based neighborhoods $N_{uu}$, $N_{uv}$, $N_{vv}$, and $N_{vu}$. It is noteworthy that the embedding process of the two types of node is performed iteratively, which means the updated vector $h_{k + 1}^u$ of worker nodes are leveraged to calculate the neighbor information within the neighborhood $N_{vu}$. After $K$ rounds embedding with identical process but independent trainable parameters, the finial vectors of worker node and subtask node can be represented as $h_{K}^u$ and $h_{K}^v$.

\subsubsection{Graph Embedding}
The network architecture of reinforcement learning is typically predetermined, necessitating the input of state and action within this fixed structure. To acquire the state embedding, we employ mean pooling to aggregate each type of node vector before concatenating these vectors together, as follows:

\begin{equation}
\label{eq.17}
{h^s} = {\mathop{\rm cat}\nolimits} \left( {{1 \over n}\sum\limits_{u \in {\cal U}} {h_K^u} ,{1 \over {ml}}\sum\limits_{v \in {\cal V}} {h_K^v} } \right)
\end{equation}

Besides, the embedding of the action $a = (u,v)$ can be obtained by concatenating the vector of the worker and the subtask, as well as the state, as follows:

\begin{equation}
\label{eq.18}
{h^a} = {\mathop{\rm cat}\nolimits} \left( {h_K^u,h_K^v,{h^s}} \right)
\end{equation}

\subsection{Decision Making and Training}

The action space encompasses all possible combinations of workers and subtasks. However, there may exist infeasible actions that cannot be selected at each step of the MDP. Hence, by employing the technique of invalid action masking\cite{xu2023intelligent}, we can effectively eliminate these infeasible actions from the action space during decision-making and only assign incomplete subtasks to available workers. 

The decision-making process utilizes a Multi-Layer Perception (MLP) as the policy network to determine the probability of selection for each feasible action, as follow:

\begin{equation}
\pi (a|s) = {{\exp ({{{\mathop{\rm MLP}\nolimits} }_\pi }({h^a}))} \over {\sum\nolimits_{\bar a \in \bar A} {\exp ({{{\mathop{\rm MLP}\nolimits} }_\pi }({h^{\bar a}}))} }}\forall a \in \bar A
\end{equation}
where $\bar A$ is the set of all feasible actions and ${{\mathop{\rm MLP}\nolimits} _\pi }(\centerdot)$ has two ${\lambda _\pi }$-dimensional hidden layers. To facilitate exploration, the action is sampled according to a probability distribution during the training phase. However, during validation and testing, the action is selected greedily based on the maximum probability.

The policy network and CHANet are trained simultaneously using the PPO method, which is a popular on-policy DRL method with the actor–critic structure. The training process follows the previous work \cite{song2022flexible}, as shown in Algorithm \ref{a2}. The training process is conducted for ${\cal I}$ iterations, with the training batch $\cal B$ being replaced every 20 episodes and the validation being performed every 10 episodes. 

\begin{algorithm}[tp]
	\caption{Training process based on PPO.}
	\hspace*{0.02in} {\bf Input:} 
       CHANet, policy network, and critic network \\
	\hspace*{0.02in} {\bf Output:} %算法的结果输出
	the trained network
	\begin{algorithmic}[1]
         \State Generate a batch of instances $\cal B$ of DMA problem 
		\For {$iter = 1:\cal I$}
                \State Initialize the state of all instances in $\cal B$
                \While{state of instances in $\cal B$ are not terminal}
                    \State Extract embeddings using CHANet
                    \State Sample action using policy network 
                    \State Conduct the action and receive reward
                    \State Transit to the next state
                \EndWhile
                \State Compute the generalisd advantage estimation

        \State Compute the PPO loss and update the network
        \If{$\bmod (iter, 10) =  = 0$}
            \State Validate the current policy
        \EndIf
        \If{$\bmod (iter, 20) =  = 0$}
            \State Generate a new batch of instances $\cal B$
        \EndIf
		\EndFor
        \State \textbf{Return}
	\end{algorithmic}
 \label{a2}
\end{algorithm}

\section{Evaluation}

We conduct extensive experiments in this section to answer the following research questions:

\begin{itemize}
    \item \textbf{RQ1}: Can the proposed HGRL-TA address the DMA problem effectively, compared with the state-of-the-art methods?
    
    \item \textbf{RQ2}: How does the proposed CHANet performed across various instances with different structures and scales?

    \item \textbf{RQ3}: What are the effects of the dependency and skill matching constraints on the performance of the proposed CHANet?
    
\end{itemize}

In what follows, the experimental settings are first described, and then answers of the above three research questions are presented.

\subsection{Baselines and Time Complexity}
This paper employs five baselines for comparison, encompassing a heuristic approach, a meta-heuristic algorithm, and three HGNNs with distinct architectures. The three HGNNs are employed within the HGRL-TA framework and trained using the PPO algorithm under identical conditions as CHANet.

\begin{itemize}

\item \textbf{DMA-G}. It utilizes the greedy policy to select the feasible action with maximum profit at each step until all valid assignments are exhausted.

\item \textbf{2SGA}. It is a two-stage genetic algorithm \cite{defersha2020efficient} specifically designed to address the Flexible Job-shop Scheduling Problem (FJSP), which involves allocating a set of operations to multiple machines and necessitates consideration of dependency relations among these operations, resembling the DMA problem. The original 2SGA is adapted to solve the DMA problem, including modifying the objective function and the state transition process.

\item \textbf{HGNN-F}. It is a type of HGNN specifically designed for addressing the FJSP, which also employs two-stage embedding to acquire the final representation of machine nodes and operation nodes \cite{song2022flexible}. However, the GAT is exclusively employed for the embedding of machine nodes, while a MLP is utilized for operation embedding to aggregate the feature of neighboring machine nodes and dependent operation nodes. It can be adapted to solve the DMA problem by mapping workers to machines and subtasks to operations.

\item\textbf{HGAT-MB}. It employs a meta-path-based method to integrate neighboring features and semantic features, as shown in Fig. \ref{fig:cp}.

\item\textbf{HGAT-MF}. It employs a meta-path-free method to integrate neighboring and semantic information simultaneously \cite{yang2023simple}.

\end{itemize}

Given the instance with $n$ workers and $ml$ subtasks, the DMA-G needs to select actions from all combinations of workers and subtasks, and the selection rounds cannot exceed $ml$ to solve the instance. Therefore, the time complexity of the DMA-G can be formulated as ${\cal O}(n \times m{l^2})$. The 2SGA is time-consuming due to the requirement of conducting comprehensive exploration through the process of population evolution. Given the population size $ps$ and the maximum generation $mg$, the time complexity of 2SGA is denoted as ${\cal O}(ps \times mg \times n \times ml)$, where $ps$ and $mg$ are set to the scale of the problem, i.e., $ps,mg \propto n \times ml$. For the GRL-based method, the time consumed by the artificial neural network to obtain the selection probability of each action can be denoted as ${W_g}$, then the time complexity of the RL-based method can be represented as  ${\cal O}(n \times m{l^2} \times {W_g})$, referring to the DMA-G. 

The ${W_g}$ primarily consists of the embedding component and decision component, with the decision component being consistent across different HGNNs examined in this paper. Therefore, the analysis of the time complexity associated with the embedding process using different HGNNs is presented herein. The HGNN-F exclusively employs the attention mechanism in worker embedding, resulting in a time complexity denoted as ${\cal O}(n \times ml)$, since the attention coefficient with each subtask needs to be calculated for every worker. Besides, the time complexity of the subtask embedding is denoted as ${\cal O}(ml)$. Therefore, the overall time complexity for HGNN-F can be represented as ${\cal O}(n \times ml){\rm{ + }}{\cal O}(ml){\rm{ = }}{\cal O}(n \times ml)$. The time complexity of HGAT-MB in worker embedding and subtask embedding can be represented as ${\cal O}(n \times (ml + ml + n))$ and ${\cal O}(ml \times (ml + ml + n + n))$ respectively, as it necessitates the aggregation of neighboring features through seven types of meta-path. Thus, the overall time complexity of HGAT-MB is denoted as ${\cal O}({n^2} + n \times ml + m{l^2}))$. Moreover, the time complexity of HGAT-MF is denoted as ${\cal O}(n \times ml)$. For the proposed CHANet, its overall time complexity can be denoted as ${\cal O}({n^2} + n \times ml + m{l^2}))$, which is consistent with that of HGAT-MB.

\subsection{Experiment Scenarios}

We use the synthetic data to test our proposed method and the parameters are presented in Table \ref{tab:II}. The spatial crowdsourcing campaign area is defined as a square with a length of 10 units. Consequently, the workers' and subtasks' locations are uniformly generated within the range of [0, 10]. Besides, the arrive time, work time, and speed of each worker are generated from the uniform distribution with range [0, 30], [20, 30], and [1, 3] respectively. The total skill pool consists of 4 types of skill, from which a random subset is uniformly selected for each worker and subtask. The maximum number of skills per worker and subtask is limited to 3. Moreover, the budget of each subtask is selected uniformly in the range [2, 5]. The number of subtasks in each task is assumed to be generated uniformly from the range [3, 5]. For subtasks belonging to the same task, they share a deadline that follows a uniform distribution within the range [40, 60]. Besides, the execution time of all subtasks are set as 1.

\begin{figure}
	\centering
	\includegraphics[width=0.75\linewidth]{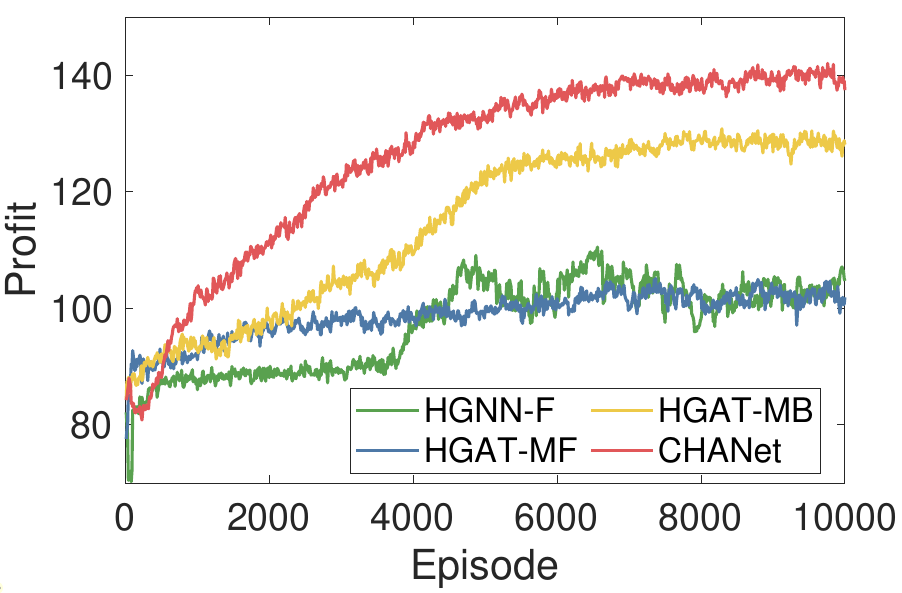}	\caption{The training curves of different HGNNs.}
	\label{fig:training}
\end{figure}

\begin{table}
\centering
\caption{Parameters of the experiment scenarios.}
\label{tab:II}
\renewcommand\arraystretch{1.2}
\resizebox{\linewidth}{!}{
\begin{tabular}{lp{1.5cm} l}
\hline
\hline
\textbf{Parameters}   & \textbf{Value} \\
\hline
size of campaign area  & $10 \times 10$ \\
arrive time of worker  & [0, 30]          \\
work time of worker    & [20, 30]      \\
speed of worker        & [1, 3]  \\
size of skill pool     & 4         \\
maximum skill number of worker and subtask  & 3    \\
budget of subtask     & [2, 5]        \\
number of subtasks in a task   & [3, 5]   \\
deadline of task     & [40, 60]    \\
\hline
\hline
\end{tabular}}
\end{table}

\subsection{Hyperparameters and Training Process}

The main hyperparameters for the proposed CHANet are presented herein, with reference to the previous work \cite{song2022flexible}. The embedding round of the CHANet is set as $K = 4$, and the embedding dimension is defined as $\lambda {\rm{ = }}16$. Besides, for the PPO method, the dimension of hidden layers in ${{\mathop{\rm MLP}\nolimits} _\pi }(\centerdot)$ is specified as ${\lambda _\pi }{\rm{ = 128}}$, and the PPO optimization epoch is determined as 3. The coefficients for the policy loss (with a clip of 0.2), value loss, and entropy term in the PPO loss function are set to 1, 0.5, and 0.01 respectively. The discount factor is set to 1.0, and network updates are performed using the Adam optimizer. The value of $\alpha $ in Eq. \ref{eq:reward} is set as 0.4. The training process consists of 10000 iterations, and each training batch contains 20 instances of the DMA problem. The policy is validated on a independent validation set that consists of 100 instances. These hyperparameters are also employed for the three baseline HGNNs.

\begin{figure}
	\centering
	\includegraphics[width=\linewidth]{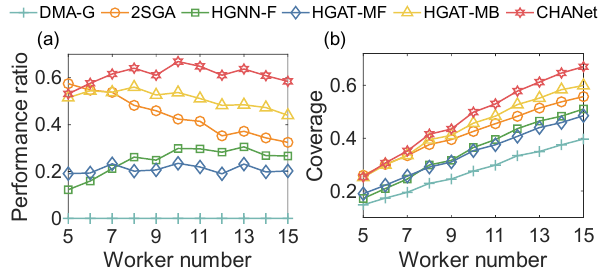}
	\caption{Performance with varying number of workers.}
	\label{fig:ep1}
\end{figure}

The training dataset and validation dataset are generated based on the parameters specified in Table \ref{tab:II}, while the number of workers and tasks is set to 10 and 20 respectively. Referring to Table \ref{tab:II}, each task consists of a range of 3-5 subtasks, thus resulting in an indeterminate total number of nodes within an instance. Therefore, the training dataset is generated during the training process, while the validation dataset is generated prior to training and utilized for the training process of various HGNNs.  The training and subsequent experiments are conducted on a machine equipped with an AMD Ryzen 7 5800 CPU and Nvidia RTX3060 GPU. The code is implemented using Python 3.9. \footnote{ The code will be provided after publication of this paper}

The training curves of CHANet and the three HGNNs are show in Fig. \ref{fig:training}, which are witnessed the convergence trends after 10000 training episodes.

The performance of the proposed CHANet in this paper stabilizes after 6000 training episodes, converging to a relatively high value can be observed. In comparison, the performance of the other three baseline HGNNs is found to be inferior to that of CHANet after convergence. Moreover, it is worth noting that the convergence speed of HGAT-MB is slower than that of CHANet, thereby demonstrating the effectiveness of our designed compound-path-based method.  The training duration for CHANet was approximately 4.3 hours, slightly shorter than the 4.7 hours required by HGAT-MB. HGNN-F and HGAT-MF utilized a more streamlined node embedding structure, resulting in reduced training times of around 3.5 hours for both networks.

\subsection{Performance Comparison (RQ1 \& RQ2)}

We assess the performance of the trained models and baselines in scenarios with  diverse structures and scales. Specifically, three groups of experiment are conducted. In the first group, we kept the task number constant at 20 while varying the number of workers from 5 to 15. In the second group, we maintained a constant worker count of 10 and varied the number of tasks from 10 to 30. For the third group, we expanded our instances by increasing both worker counts (ranging between 20 and 40) and task numbers (ranging between 30 and 50). The other parameters are set to their default values as shown in Table \ref{tab:II}. To present a comprehensive evaluation, 100 instances are generated for each setup in the three experiment groups. Besides, two metrics are utilized for performance evaluation. The first metric is the performance ratio, which calculates the profit increase ratio of other schemes compared to the DMA-G, serving as a benchmark. Another indicator is subtask coverage, representing the proportion of completed subtasks out of the total number.

\subsubsection{Performance with Varying Number of Workers} 

As shown in Fig. \ref{fig:ep1}(a), when the number of workers is five, the problem size is relatively small, and the 2SGA algorithm achieves optimal performance.  However, as the problem size increases with the number of workers, the performance of the 2SGA algorithm consistently declines.  In addition, the performance of HGAT-MB also shows a slight downward trend as the number of workers increases.  In contrast, CHANet maintains a stable and high level of performance across different scenarios with varying numbers of workers, demonstrating the network's good generalisation capabilities.  Moreover, the performance of HGNN-F and HGAT-MF remains at a lower level but still surpasses that of the DMA-G algorithm.  This indicates the effectiveness of the GRL method, although different network structures significantly impact its performance.  Fig. \ref{fig:ep1}(b) illustrates the subtask coverage of these methods in various scenarios, where the subtask coverage gradually increases with the number of workers, and CHANet's performance advantage in coverage becomes increasingly evident.

\subsubsection{Performance with Varying Number of Tasks}

\begin{figure}
	\centering
	\includegraphics[width=\linewidth]{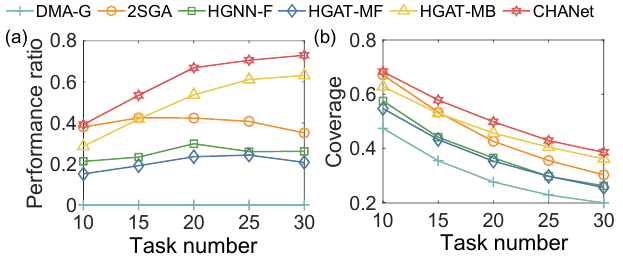}
	\caption{Performance with varying number of tasks.}
	\label{fig:ep2}
\end{figure}

The experimental results depicted in Fig. \ref{fig:ep2} exhibit similarities to those illustrated in Fig. \ref{fig:ep1}.   When the number of tasks is limited, it corresponds to a smaller problem size, thereby  the 2SGA algorithm can obtain the optimal performance.  However, as the number of tasks increases, the performance of the 2SGA algorithm gradually deteriorates. In contrast, both CHANet and HGAT-MB exhibit a clear upward trend in performance with increasing task numbers, with CHANet consistently outperforming HGAT-MB. This advantage in performance arises due to the larger solution space available for task allocation problem when there are more tasks, enabling CHANet to identify superior allocation strategies. As the number of tasks increases from 20 to 30, the upward trend gradually decelerates due to the limited working hours of the employees, impeding them from completing additional subtasks and thus hindering their ability to achieve greater profits. The performance of HGNN-F and HGAT-MF remains consistently low across scenarios with various task numbers, with a slight advantage for HGNN-F over HGAT-MF. This could be attributed to the utilization of a two-stage embedding approach in the HGNN-F, which facilitates the transmission of node features through longer paths.

The coverage of the six schemes under different task numbers is illustrated in Fig. \ref{fig:ep2}(b). As the number of tasks increases, so does the number of subtasks. However, due to workers' limited work time, they are unable to cover an increasing number of subtasks, resulting in a gradual decrease in subtask coverage. Despite the decrease in coverage as the number of tasks increases, CHANet consistently achieves the highest coverage across all scenarios, showcasing its robust generalization capabilities across varying task numbers.

\subsubsection{Performance on Large-scale Instances}

\begin{figure*}[!htb]
	\centering
    \includegraphics[width=0.9\textwidth]{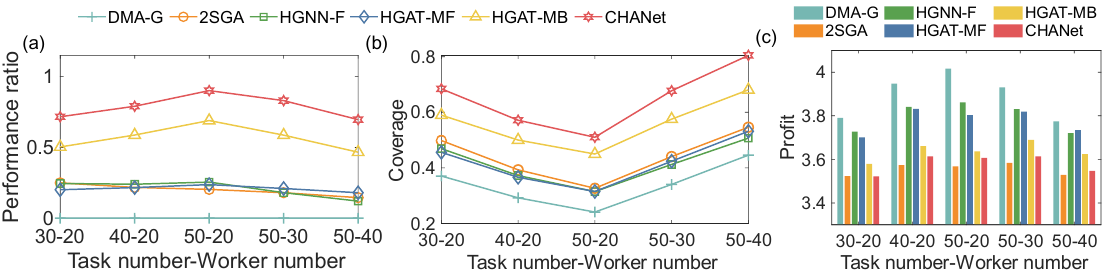}
    \caption{Performance on large-scale instances.} 
	\label{fig:ep3}
\end{figure*}

The results are presented in Fig. \ref{fig:ep3}, where CHANet continues to exhibit robust performance on the large-scale test set with an average performance ratio of 0.7870 across five scenarios, surpassing that of HGAT-MB at 0.5667. Conversely, the performances of 2SGA, HGNN-F and HGAT-MF remained suboptimal. Notably, in some scenarios, 2SGA underperformed compared to HGNN-F and HGAT-MF; however, Fig. \ref{fig:ep1} and Fig. \ref{fig:ep2} consistently demonstrate superior performance by this algorithm over both HGNN-F and HGAT-MF.

The coverage of six schemes is illustrated in Fig. \ref{fig:ep3} (b), indicating that schemes with a higher performance ratio generally exhibit higher coverage. The proposed CHANet achieves high coverage across all scenarios.  However, for the 2SGA, although its performance falls short of HGNN-F and HGAT-MF in certain scenarios, its coverage consistently surpasses that of these two methods. This suggests that the average profit from the subtasks completed by this method is lower than those of the other two. Additionally, Fig. \ref{fig:ep3} (c) presents the average profit of completed subtasks across all schemes. Based on the parameters in Table \ref{tab:II}, the expected average profit for all subtasks is 3.5. However, it is noteworthy that the DMA-G, HGNN-F, and HGAT-MF schemes outperform this average in terms of their subtask profits, while the remaining three schemes fall short. The DMA-G algorithm attains the highest average profits for completed subtasks owing to its inherent greediness. In parallel, HGNN-F and HGAT-MF also yield substantial average profits, implying that their learned policies may exhibit resemblances to greedy strategies, albeit with certain distinctions, as both approaches generate total profits surpassing those of DMA-G. Moreover, HGAT-MB and CHANet demonstrate lower average subtask profits due to the trade-off between achieving higher total profits and completing less profitable prerequisite subtasks.

\subsubsection{Running Time Analysis}

\begin{figure*}[!htb]
	\centering
    \includegraphics[width=0.9\textwidth]{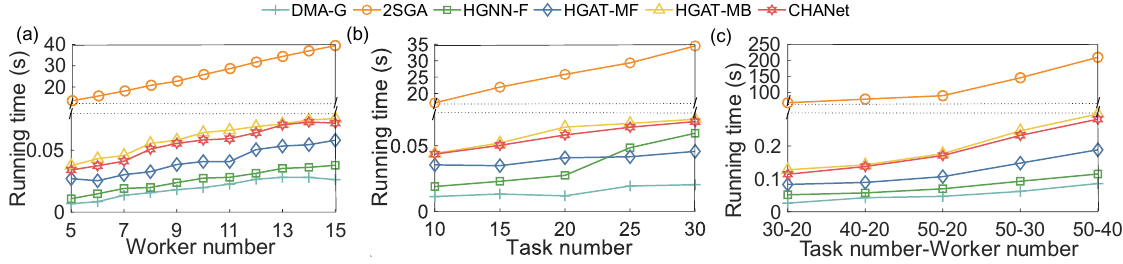}
    \caption{Running Time Analysis.} 
	\label{fig:eptime}
\end{figure*}

The previous section introduced the time complexities of the six schemes. This section presents the average running time of these schemes across different scenarios, where average running time refers to the mean time taken by each scheme to obtain a solution for a given instance, as depicted in Fig. \ref{fig:eptime}. It is evident that the running time of all schemes increases proportionally with the growth of problem size. The DMA-G scheme consistently demonstrates the shortest running time across all scenarios, and its increase in running time with problem size is relatively gradual due to its comparatively straightforward decision-making process. The running time of the 2SGA algorithm, in contrast to other schemes, significantly exceeds particularly in scenarios with larger problem sizes.  This is attributed to its requirement for extensive exploration of the solution space in order to continuously optimize the solution, resulting in a rapid increase in running time as the problem size expands. For GRL-based methods, the running time also exhibits a gradual increase with problem size.  Specifically, HGNN-F and HGAT-MF demonstrate slightly higher running times compared to DMA-G, yet lower than HGAT-MB and CHANet.  The running times of HGAT-MB and CHANet are comparable, with CHANet generally showcasing marginally lower times than HGAT-MB, which aligns with their respective training time rankings.

To summarize, the proposed HGRL-TA demonstrates effective resolution of the DMA problem in comparison to the heuristic method (DMA-G) and the metaheuristic method (2SGA). It achieves average profits across all scenarios that are 65.23\% and 21.78\% higher than those obtained using the DMA-G and the 2SGA, respectively. Furthermore, it offers a significant advantage in terms of computational efficiency compared to the 2SGA.  However, the performance of HGRL-TA is heavily reliant on the well-designed CHANet, which demonstrates robust generality across instances with diverse structures and scales, even on those instances that are dissimilar to the training set.

\subsection{Impact of Constraints (RQ3)}

\begin{figure}
	\centering
	\includegraphics[width=\linewidth]{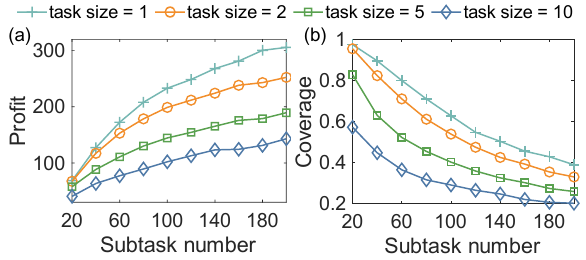}
	\caption{Performance with varying subtask number and task size.}
	\label{fig:ep4}
\end{figure}

\begin{figure*}[!htb]
	\centering
	\includegraphics[width=\textwidth]{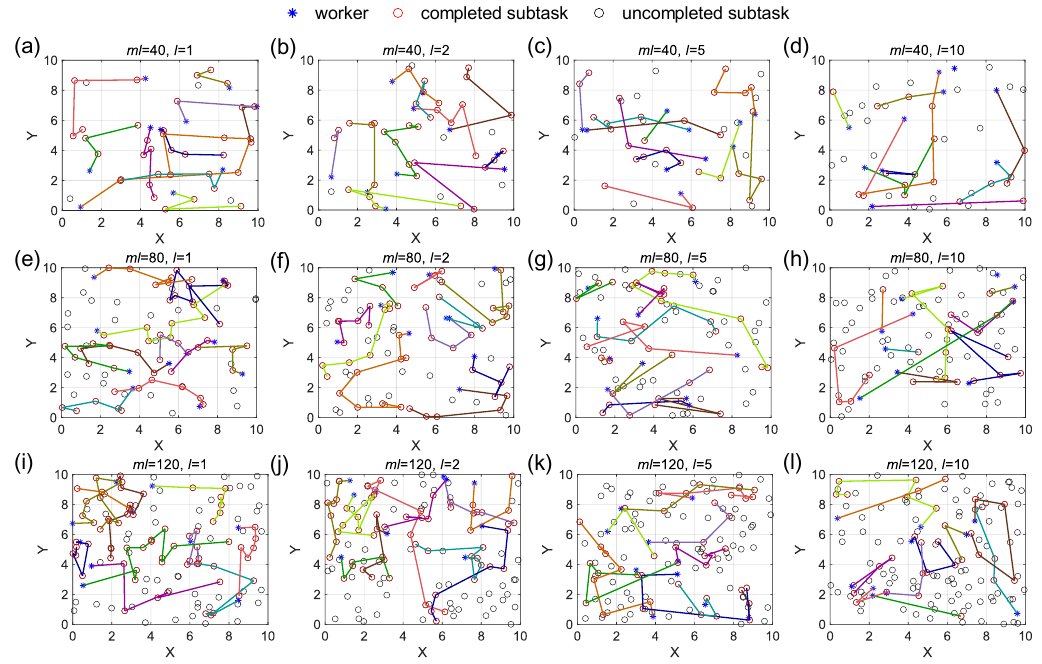}
	\caption{The subtask allocation results.}
	\label{fig:trajectory}
\end{figure*}

This section investigates the impact of dependency and skill matching constraints on CHANet performance by manipulating their associated parameters. Firstly, we examine the influence of dependency constraints through variations in both the total number of subtasks $ml$ and task sizes $l$. Subsequently, we explore the effects of skill matching constraints by altering the maximum skill numbers possessed by workers and required for subtasks.

\subsubsection{Impact of Dependency Constraint}

The dependency constraints in this paper exist between subtasks within the same task. In previous experiments, the number of tasks was varied, but the task size, meaning the number of subtasks included in each task, was randomly generated within a range of [3,5]. To thoroughly explore the impact of the dependency constraints, we first generated a certain number of subtasks and then divided them into tasks of different sizes to ensure that the total budget in the instances remains consistent level. Specifically, we set the total number of subtasks $ml$ to be between 20 and 200, with task sizes $l$ ranging from 1 to 10, while the other parameters adopted the default values from Table \ref{tab:II}. The experimental results are shown in Fig. \ref{fig:ep4}.

The smaller the task size, the fewer dependencies exist between subtasks. When the task size equals 1, there are no dependency among subtasks, and at this point, the DMA problem is reduced to a simple multi-task allocation problem. The results depicted in Fig. \ref{fig:ep4} demonstrate that CHANet achieves maximum profit across all scenarios with varying total numbers of subtasks when the task size equals 1. However, as the task size increases, there is a gradual decline in profitability.  The performance gap between different task sizes widens with an increasing number of subtasks.  When the number of subtasks increases, coverage exhibits a downward trend, irrespective of the presence of dependency constraints.  With equal subtask number, large task sizes result in low coverage, indicating complete completion of few subtasks and consequently yielding small profits.

 The experimental results above demonstrate that dependency constraints have a significant impact on the task allocation. To further analyze the underlying mechanisms, we present the allocation schemes obtained by CHANet in instances with varying numbers of subtasks and task sizes, as illustrated in Fig. \ref{fig:trajectory}. It is evident that, when considering larger task sizes, fewer subtasks can be completed under the same number of subtasks. For instance, the number of completed subtasks in (a), (b), (c), and (d) are 36, 34, 27, and 18, respectively. This is because the dependencies between subtasks reduce the number of valid worker-subtask pairs at each step. Consequently, workers require more time to reach the locations of these subtasks, thereby limiting their ability to complete a higher number of tasks within a given work time. The solution space expands as the number of subtasks increases, providing workers with a wider range of profitable options at each step. This enables them to achieve higher profits within the limited work time. To demonstrate that, the average profit per unit of time for the workers can be calculated by dividing the total profits by the cumulative movement time of all workers. The values in (a), (e), and (i) are 0.42, 0.96, and 1.06 respectively, indicating a clear upward trend as the number of subtasks increases. However, as the number of subtasks increases further, it becomes increasingly challenging for workers to generate additional profits within the limited time. Consequently, the trend of profit growth gradually decelerates, which is also observed in Fig. \ref{fig:ep4}.

\subsubsection{Impact of Skill Matching Constraint}

\begin{figure}
	\centering
	\includegraphics[width=\linewidth]{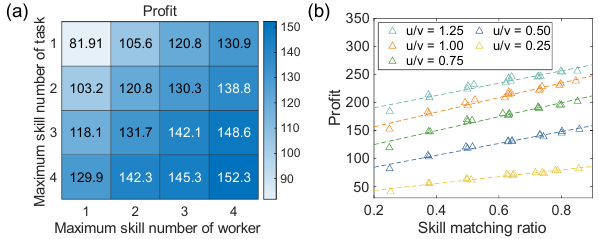}
	\caption{Performance of CHANet with varying number of skills per worker and subtask.}
	\label{fig:ep5}
\end{figure}

The training set allows each worker to acquire up to three distinct skills, while each subtask can be accomplished using a maximum of three different skills.  In order to investigate the impact of skill matching constraint, we modify these two parameters accordingly.  The number of workers and tasks are maintained consistent with those in the training set and then adjust the maximum number of skills for both workers and subtasks within a range from 1 to 4, considering that there are a total of four available skills. The results are shown in Fig. \ref{fig:ep5} (a). The increase in the maximum number of skills for workers and subtasks is observed to positively correlate with the overall profits.  This can be attributed to the fact that a higher maximum skill number enhances the likelihood of skill matching between workers and subtasks, thereby providing workers with a wider range of options to achieve greater profitability.

To further investigate the correlation between profit growth and the number of skills, the skill matching ratio is determined by calculating the proportion of worker-subtask pairs that exhibit skill matching relationships across scenarios with varying maximum skill numbers. Subsequently, the skill matching ratios are plotted as the x-coordinate, and the corresponding profits for each scenario are plotted as the y-coordinate. The data is projected onto a two-dimensional plane and a linear fit is performed, as illustrated in Fig. \ref{fig:ep5} (b). For a comprehensive comparison, we held the number of tasks constant at 20 while varying the number of workers from 5 to 25 in order to evaluate CHANet's performance under scenarios with different worker-to-task ratio $u/v$. The relationship between the increase in profits and the rise in skill matching ratio across various $u/v$ values demonstrates a nearly linear correlation, exhibiting similar growth rates. At the same skill matching ratio, a higher $u/v$ results in greater achievable profits, consistent with conclusions drawn from previous experiments.

\section{Conclusion and Future Work}

This paper aims to address a specific task allocation problem in spatial crowdsourcing, known as the DMA problem. To this end, we propose a HGRL-TA framework. Within this framework, we represent the problem state using a multi-relation graph model and construct edges in the graph based on three types of relationships between nodes: skill matching, dependent, and adjacent. Moreover, we design a CHANet to effectively embed the problem states. The experimental results confirm the superior performance of our proposed HGRL-TA in comparison with DMA-G and 2SGA. Furthermore, the Compound-path-based method (CHANet) outperforms the Meta-path-based method (HGAT-MB) due to its ability to aggregate features based on comprehensive node relationships, resulting in more suitable node embeddings for the task allocation problem. The DMA problem addressed in this paper, however, is offline and does not fully exploit the advantages of HGRL methods. Future research could further investigate solutions for online DMA problems.

\bibliographystyle{./IEEEtran}
\bibliography{tmc_reference}

\begin{thebibliography}{10}
\providecommand{\url}[1]{#1}
\csname url@samestyle\endcsname
\providecommand{\newblock}{\relax}
\providecommand{\bibinfo}[2]{#2}
\providecommand{\BIBentrySTDinterwordspacing}{\spaceskip=0pt\relax}
\providecommand{\BIBentryALTinterwordstretchfactor}{4}
\providecommand{\BIBentryALTinterwordspacing}{\spaceskip=\fontdimen2\font plus
\BIBentryALTinterwordstretchfactor\fontdimen3\font minus \fontdimen4\font\relax}
\providecommand{\BIBforeignlanguage}[2]{{%
\expandafter\ifx\csname l@#1\endcsname\relax
\typeout{** WARNING: IEEEtran.bst: No hyphenation pattern has been}%
\typeout{** loaded for the language `#1'. Using the pattern for}%
\typeout{** the default language instead.}%
\else
\language=\csname l@#1\endcsname
\fi
#2}}
\providecommand{\BIBdecl}{\relax}
\BIBdecl

\bibitem{tong2020spatial}
Y.~Tong, Z.~Zhou, Y.~Zeng, L.~Chen, and C.~Shahabi, ``Spatial crowdsourcing: a survey,'' \emph{The VLDB Journal}, vol.~29, pp. 217--250, 2020.

\bibitem{liu2020crowdos}
Y.~Liu, Z.~Yu, B.~Guo, Q.~Han, J.~Su, and J.~Liao, ``Crowdos: A ubiquitous operating system for crowdsourcing and mobile crowd sensing,'' \emph{IEEE Transactions on Mobile Computing}, vol.~21, no.~3, pp. 878--894, 2020.

\bibitem{wang2020compact}
L.~Wang, Z.~Yu, Q.~Han, D.~Yang, S.~Pan, Y.~Yao, and D.~Zhang, ``Compact scheduling for task graph oriented mobile crowdsourcing,'' \emph{IEEE Transactions on Mobile Computing}, vol.~21, no.~7, pp. 2358--2371, 2020.

\bibitem{zhang2022online}
J.~Zhang, T.~Jiang, X.~Gao, and G.~Chen, ``An online fairness-aware task planning approach for spatial crowdsourcing,'' \emph{IEEE Transactions on Mobile Computing}, vol.~23, no.~1, pp. 150--163, 2022.

\bibitem{xu2022incentive}
Y.~Xu, M.~Xiao, J.~Wu, S.~Zhang, and G.~Gao, ``Incentive mechanism for spatial crowdsourcing with unknown social-aware workers: A three-stage stackelberg game approach,'' \emph{IEEE Transactions on Mobile Computing}, vol.~22, no.~8, pp. 4698--4713, 2022.

\bibitem{wang2023ropriv}
M.~Wang, H.~Jiang, P.~Zhao, J.~Li, J.~Liu, G.~Min, and S.~Dustdar, ``Ropriv: Road network-aware privacy-preserving framework in spatial crowdsourcing,'' \emph{IEEE Transactions on Mobile Computing}, vol.~23, no.~3, pp. 2351--2366, 2023.

\bibitem{estelles2012towards}
E.~Estell{\'e}s-Arolas and F.~Gonz{\'a}lez-Ladr{\'o}n-de Guevara, ``Towards an integrated crowdsourcing definition,'' \emph{Journal of Information science}, vol.~38, no.~2, pp. 189--200, 2012.

\bibitem{liu2018foodnet}
Y.~Liu, B.~Guo, C.~Chen, H.~Du, Z.~Yu, D.~Zhang, and H.~Ma, ``Foodnet: Toward an optimized food delivery network based on spatial crowdsourcing,'' \emph{IEEE Transactions on Mobile Computing}, vol.~18, no.~6, pp. 1288--1301, 2018.

\bibitem{li2022auction}
Y.~Li, Y.~Li, Y.~Peng, X.~Fu, J.~Xu, and M.~Xu, ``Auction-based crowdsourced first and last mile logistics,'' \emph{IEEE Transactions on Mobile Computing}, vol.~23, no.~1, pp. 180--193, 2022.

\bibitem{wang2014smartphoto}
Y.~Wang, W.~Hu, Y.~Wu, and G.~Cao, ``Smartphoto: a resource-aware crowdsourcing approach for image sensing with smartphones,'' in \emph{Proceedings of the 15th ACM international symposium on mobile ad hoc networking and computing}, 2014, pp. 113--122.

\bibitem{wang2023leto}
Y.~Wang, A.~K.-S. Wong, S.-H.~G. Chan, and W.~H. Mow, ``Leto: crowdsourced radio map construction with learned topology and a few landmarks,'' \emph{IEEE Transactions on Mobile Computing}, vol.~23, no.~4, pp. 2795--2812, 2023.

\bibitem{liu2022multi}
Z.~Liu, K.~Li, X.~Zhou, N.~Zhu, Y.~Gao, and K.~Li, ``Multi-stage complex task assignment in spatial crowdsourcing,'' \emph{Information Sciences}, vol. 586, pp. 119--139, 2022.

\bibitem{yao2022online}
J.~Yao, L.~Yang, and X.~Xu, ``Online dependent task assignment in preference aware spatial crowdsourcing,'' \emph{IEEE Transactions on Services Computing}, vol.~16, no.~4, pp. 2827--2840, 2022.

\bibitem{ni2020task}
W.~Ni, P.~Cheng, L.~Chen, and X.~Lin, ``Task allocation in dependency-aware spatial crowdsourcing,'' in \emph{2020 IEEE 36th International Conference on Data Engineering (ICDE)}.\hskip 1em plus 0.5em minus 0.4em\relax IEEE, 2020, pp. 985--996.

\bibitem{cheng2016task}
P.~Cheng, X.~Lian, L.~Chen, J.~Han, and J.~Zhao, ``Task assignment on multi-skill oriented spatial crowdsourcing,'' \emph{IEEE Transactions on Knowledge and Data Engineering}, vol.~28, no.~8, pp. 2201--2215, 2016.

\bibitem{gao2016top}
D.~Gao, Y.~Tong, J.~She, T.~Song, L.~Chen, and K.~Xu, ``Top-k team recommendation in spatial crowdsourcing,'' in \emph{International conference on web-age information management}.\hskip 1em plus 0.5em minus 0.4em\relax Springer, 2016, pp. 191--204.

\bibitem{li2019multi}
X.~Li and X.~Zhang, ``Multi-task allocation under time constraints in mobile crowdsensing,'' \emph{IEEE Transactions on Mobile Computing}, vol.~20, no.~4, pp. 1494--1510, 2019.

\bibitem{liu2016taskme}
Y.~Liu, B.~Guo, Y.~Wang, W.~Wu, Z.~Yu, and D.~Zhang, ``Taskme: Multi-task allocation in mobile crowd sensing,'' in \emph{Proceedings of the 2016 ACM international joint conference on pervasive and ubiquitous computing}, 2016, pp. 403--414.

\bibitem{lu2023incentivizing}
J.~Lu, H.~Liu, R.~Jia, Z.~Zhang, X.~Wang, and J.~Wang, ``Incentivizing proportional fairness for multi-task allocation in crowdsensing,'' \emph{IEEE Transactions on Services Computing}, 2023.

\bibitem{shen2023heterogeneous}
X.~Shen, D.~Xu, L.~Song, and Y.~Zhang, ``Heterogeneous multi-project multi-task allocation in mobile crowdsensing using an ensemble fireworks algorithm,'' \emph{Applied Soft Computing}, vol. 145, p. 110571, 2023.

\bibitem{han2021online}
L.~Han, Z.~Yu, Z.~Yu, L.~Wang, H.~Yin, and B.~Guo, ``Online organizing large-scale heterogeneous tasks and multi-skilled participants in mobile crowdsensing,'' \emph{IEEE Transactions on Mobile Computing}, vol.~22, no.~5, pp. 2892--2909, 2021.

\bibitem{zhu2020cost}
Z.~Zhu, B.~Chen, W.~Liu, Y.~Zhao, Z.~Liu, and Z.~Zhao, ``A cost-quality beneficial cell selection approach for sparse mobile crowdsensing with diverse sensing costs,'' \emph{IEEE Internet of Things Journal}, vol.~8, no.~5, pp. 3831--3850, 2020.

\bibitem{zhu2022crowd}
Z.~Zhu, Y.~Zhao, B.~Chen, S.~Qiu, Z.~Liu, K.~Xie, and L.~Ma, ``A crowd-aided vehicular hybrid sensing framework for intelligent transportation systems,'' \emph{IEEE Transactions on Intelligent Vehicles}, vol.~8, no.~2, pp. 1484--1497, 2022.

\bibitem{zhao2023cost}
Y.~Zhao, Z.~Zhu, and B.~Chen, ``Cost-quality aware compressive mobile crowdsensing,'' in \emph{Mobile Crowdsourcing: From Theory to Practice}.\hskip 1em plus 0.5em minus 0.4em\relax Springer, 2023, pp. 225--247.

\bibitem{tao2020profit}
X.~Tao and W.~Song, ``Profit-oriented task allocation for mobile crowdsensing with worker dynamics: Cooperative offline solution and predictive online solution,'' \emph{IEEE Transactions on Mobile Computing}, vol.~20, no.~8, pp. 2637--2653, 2020.

\bibitem{peng2021graph}
Y.~Peng, B.~Choi, and J.~Xu, ``Graph learning for combinatorial optimization: a survey of state-of-the-art,'' \emph{Data Science and Engineering}, vol.~6, no.~2, pp. 119--141, 2021.

\bibitem{song2022flexible}
W.~Song, X.~Chen, Q.~Li, and Z.~Cao, ``Flexible job-shop scheduling via graph neural network and deep reinforcement learning,'' \emph{IEEE Transactions on Industrial Informatics}, vol.~19, no.~2, pp. 1600--1610, 2022.

\bibitem{zhao2024application}
Y.~Zhao, X.~Luo, and Y.~Zhang, ``The application of heterogeneous graph neural network and deep reinforcement learning in hybrid flow shop scheduling problem,'' \emph{Computers \& Industrial Engineering}, vol. 187, p. 109802, 2024.

\bibitem{guo2018task}
B.~Guo, Y.~Liu, L.~Wang, V.~O. Li, J.~C. Lam, and Z.~Yu, ``Task allocation in spatial crowdsourcing: Current state and future directions,'' \emph{IEEE Internet of Things Journal}, vol.~5, no.~3, pp. 1749--1764, 2018.

\bibitem{wang2017multi}
L.~Wang, Z.~Yu, Q.~Han, B.~Guo, and H.~Xiong, ``Multi-objective optimization based allocation of heterogeneous spatial crowdsourcing tasks,'' \emph{IEEE Transactions on Mobile Computing}, vol.~17, no.~7, pp. 1637--1650, 2017.

\bibitem{wang2022triple}
W.~Wang, Y.~Wang, P.~Duan, T.~Liu, X.~Tong, and Z.~Cai, ``A triple real-time trajectory privacy protection mechanism based on edge computing and blockchain in mobile crowdsourcing,'' \emph{IEEE Transactions on Mobile Computing}, vol.~22, no.~10, pp. 5625--5642, 2022.

\bibitem{bhatti2020approximation}
S.~S. Bhatti, J.~Fan, K.~Wang, X.~Gao, F.~Wu, and G.~Chen, ``An approximation algorithm for bounded task assignment problem in spatial crowdsourcing,'' \emph{IEEE Transactions on Mobile Computing}, vol.~20, no.~8, pp. 2536--2549, 2020.

\bibitem{zhang2019expertise}
X.~Zhang, Y.~Wu, L.~Huang, H.~Ji, and G.~Cao, ``Expertise-aware truth analysis and task allocation in mobile crowdsourcing,'' \emph{IEEE Transactions on Mobile Computing}, vol.~20, no.~3, pp. 1001--1016, 2019.

\bibitem{zhang2021multi}
J.~Zhang and X.~Zhang, ``Multi-task allocation in mobile crowd sensing with mobility prediction,'' \emph{IEEE Transactions on Mobile Computing}, vol.~22, no.~2, pp. 1081--1094, 2021.

\bibitem{estrada2017crowd}
R.~Estrada, R.~Mizouni, H.~Otrok, A.~Ouali, and J.~Bentahar, ``A crowd-sensing framework for allocation of time-constrained and location-based tasks,'' \emph{IEEE Transactions on Services Computing}, vol.~13, no.~5, pp. 769--785, 2017.

\bibitem{munikoti2023challenges}
S.~Munikoti, D.~Agarwal, L.~Das, M.~Halappanavar, and B.~Natarajan, ``Challenges and opportunities in deep reinforcement learning with graph neural networks: A comprehensive review of algorithms and applications,'' \emph{IEEE transactions on neural networks and learning systems}, 2023.

\bibitem{barrett2020exploratory}
T.~Barrett, W.~Clements, J.~Foerster, and A.~Lvovsky, ``Exploratory combinatorial optimization with reinforcement learning,'' in \emph{Proceedings of the AAAI conference on artificial intelligence}, vol.~34, no.~04, 2020, pp. 3243--3250.

\bibitem{drori2020learning}
I.~Drori, A.~Kharkar, W.~R. Sickinger, B.~Kates, Q.~Ma, S.~Ge, E.~Dolev, B.~Dietrich, D.~P. Williamson, and M.~Udell, ``Learning to solve combinatorial optimization problems on real-world graphs in linear time,'' in \emph{2020 19th IEEE International Conference on Machine Learning and Applications (ICMLA)}.\hskip 1em plus 0.5em minus 0.4em\relax IEEE, 2020, pp. 19--24.

\bibitem{almasan2022deep}
P.~Almasan, J.~Su{\'a}rez-Varela, K.~Rusek, P.~Barlet-Ros, and A.~Cabellos-Aparicio, ``Deep reinforcement learning meets graph neural networks: Exploring a routing optimization use case,'' \emph{Computer Communications}, vol. 196, pp. 184--194, 2022.

\bibitem{xu2023intelligent}
C.~Xu and W.~Song, ``Intelligent task allocation for mobile crowdsensing with graph attention network and deep reinforcement learning,'' \emph{IEEE Transactions on Network Science and Engineering}, vol.~10, no.~2, pp. 1032--1048, 2023.

\bibitem{wang2022survey}
X.~Wang, D.~Bo, C.~Shi, S.~Fan, Y.~Ye, and S.~Y. Philip, ``A survey on heterogeneous graph embedding: methods, techniques, applications and sources,'' \emph{IEEE Transactions on Big Data}, vol.~9, no.~2, pp. 415--436, 2022.

\bibitem{yang2023simple}
X.~Yang, M.~Yan, S.~Pan, X.~Ye, and D.~Fan, ``Simple and efficient heterogeneous graph neural network,'' in \emph{Proceedings of the AAAI conference on artificial intelligence}, vol.~37, no.~9, 2023, pp. 10\,816--10\,824.

\bibitem{fang2023learning}
H.~Fang, Z.~Xiao, P.~Zheng, H.~Chen, Z.~Li, J.~Bu, and H.~Wang, ``Learning co-occurrence patterns for next destination recommendation,'' \emph{IEEE Transactions on Mobile Computing}, 2023.

\bibitem{huang2023egomuil}
H.~Huang, F.~Ding, H.~Yin, G.~Liu, C.~Wang, and D.~O. Wu, ``Egomuil: Enhancing spatio-temporal user identity linkage in location-based social networks with ego-mo hypergraph,'' \emph{IEEE Transactions on Mobile Computing}, 2023.

\bibitem{zhou2023predicting}
Z.~Zhou, K.~Yang, Y.~Liang, B.~Wang, H.~Chen, and Y.~Wang, ``Predicting collective human mobility via countering spatiotemporal heterogeneity,'' \emph{IEEE Transactions on Mobile Computing}, 2023.

\bibitem{wang2019heterogeneous}
X.~Wang, H.~Ji, C.~Shi, B.~Wang, Y.~Ye, P.~Cui, and P.~S. Yu, ``Heterogeneous graph attention network,'' in \emph{The world wide web conference}, 2019, pp. 2022--2032.

\bibitem{defersha2020efficient}
F.~M. Defersha and D.~Rooyani, ``An efficient two-stage genetic algorithm for a flexible job-shop scheduling problem with sequence dependent attached/detached setup, machine release date and lag-time,'' \emph{Computers \& Industrial Engineering}, vol. 147, p. 106605, 2020.

\end{thebibliography}

\end{document}